\documentclass[11pt]{article}

\usepackage[letterpaper,margin=1in]{geometry}
\usepackage[T1]{fontenc}
\usepackage{lmodern}
\usepackage{amsmath,amssymb,amsthm}
\usepackage{graphicx,float}
\usepackage[authoryear,round]{natbib}
\usepackage{makecell,arydshln,placeins}
\usepackage{microtype}
\usepackage[font=small,labelfont=bf]{caption}
\usepackage[hidelinks,hypertexnames=false]{hyperref}
\graphicspath{{./}{../}{../results/}{paper/}{results/}}
\newtheorem{theorem}{Theorem}

\newtheorem{boundednessproposition}{Proposition}
\theoremstyle{definition}
\newtheorem{definition}[theorem]{Definition}
\newtheorem{assumption}{Assumption}
\theoremstyle{remark}

\newenvironment{keywords}
  {\par\smallskip\noindent\textbf{Keywords: }\ignorespaces}
  {\par\medskip}

\title{Graph Representation Learning via\\Adaptive Expectile Aggregation}
\author{
  \begin{minipage}{0.95\textwidth}
    \centering
    \normalsize
    Tianyi Zhu\textsuperscript{1}
    \qquad
    Wei Lan\textsuperscript{2}\thanks{Corresponding author.}
    \\[0.8em]
    \small
    \textsuperscript{1}Department of Statistics, Colorado State University\\
    Fort Collins, CO 80523, USA
    \\[0.5em]
    \textsuperscript{2}Laboratory for Data Intelligence and Innovation\\
    in Finance and Economics and Center of Statistical Research\\
    Southwestern University of Finance and Economics, Chengdu, China
    \\[0.7em]
    \texttt{tianyi.zhu@colostate.edu}
    \qquad
    \texttt{lanwei@swufe.edu.cn}
  \end{minipage}
}
\date{}

\begin{document}
\maketitle

\begin{abstract}
Graph neural networks (GNNs) rely critically on neighborhood aggregation, yet conventional mean aggregation may fail to capture class-relevant distributional information, particularly in heterophilic graphs. We propose ExpectileGCN, an adaptive graph aggregation framework that introduces a learnable expectile level to capture neighborhood information beyond the mean. The proposed operator continuously adjusts aggregation between minimum- and maximum-type behaviors, with mean aggregation as a special case. A differentiable approximation based on sigmoid smoothing and fixed-point iterations enables end-to-end optimization. Theoretically, we establish stability properties and transductive generalization guarantees for ExpectileGCN. We further investigate its classification risk under heterophilic neighborhood models. For informative neighborhoods, we show that expectile aggregation can exploit class-dependent scale differences to achieve a classification advantage over mean aggregation. Under a local asymptotic regime with diminishing per-neighbor label informativeness, we establish that this advantage can persist in terms of limiting classification risk. Experiments on homophilic and heterophilic benchmark graphs demonstrate strong predictive performance and improved robustness under feature corruption and structural perturbations.
\end{abstract}

\begin{keywords}
  Graph Neural Networks; Expectile Aggregation; Robust Graph Learning; Transductive Generalization; Neighborhood Informativeness
\end{keywords}

\section{Introduction}\label{sec-intro}
Graph Neural Networks (GNNs) learn node representations by combining node features with information from neighboring nodes \citep{kipf2017semi,hamilton2017inductive,velivckovic2018graph}. Neighborhood aggregation plays a central role in this process, as it determines how neighboring messages are summarized for downstream prediction. Many classical GNN architectures are motivated by label homophily, where connected nodes tend to share the same class labels and neighboring features provide useful information for classification \citep{mcpherson2001birds,hamilton2020graph}. However, real-world graphs often exhibit heterophily, where connected nodes may belong to different classes. This raises the question of how neighborhood information can be effectively extracted when neighboring labels do not necessarily agree with the target label.

Heterophily does not necessarily imply that neighboring nodes are uninformative. Different node classes may exhibit distinct neighborhood label distributions, allowing cross-class connections to provide useful information for classification. \citet{ma2022homophily} showed that GCNs can perform well on heterophilic graphs under suitable neighborhood distribution conditions. However, \citet{platonov2023characterizing} found that several widely used heterophilic benchmark datasets exhibit low label informativeness, indicating that the label of a neighboring node provides limited information about the target node's label. These observations motivate studying how aggregation can exploit
class-relevant information when individual neighbors provide only
weak predictive cues.

The presence of class-relevant neighborhood information does not
by itself ensure that an aggregation rule can effectively exploit it. Conventional mean aggregation summarizes neighborhood messages
through their average, which may yield poorly separated class
representations when the underlying message distributions have
similar means but differ in scale or shape. Thus, the choice of
aggregation statistic affects how distributional differences are
translated into discriminative node representations. This motivates
a learnable aggregation mechanism that can adjust its response to
different neighborhood distributions.

To address this issue, we consider expectiles, a family of location statistics defined through asymmetric least squares \citep{newey1987expectile}. Unlike the mean, expectiles assign asymmetric weights to observations above and below the aggregated value, with the degree of asymmetry controlled by a parameter $\tau\in(0,1)$. When $\tau=\frac{1}{2}$, expectile aggregation reduces to mean aggregation. Moving $\tau$ away from $\frac{1}{2}$ changes the relative influence of positive and negative deviations while retaining sensitivity to their magnitudes. Consequently, differences in distributional scale can produce distinct expectile centers even when the corresponding means coincide. With one output per feature coordinate, expectiles therefore provide a compact, continuously adjustable extension of mean aggregation that directly accommodates signed messages.

Based on these properties, we propose ExpectileGCN, a graph neural network with learnable expectile aggregation. By jointly optimizing the expectile level and network parameters, ExpectileGCN adaptively adjusts the influence of neighborhood message deviations rather than relying on a fixed aggregation statistic. To facilitate end-to-end optimization, we develop a differentiable approximation based on sigmoid smoothing and fixed-point iterations. We further establish stability properties of the resulting implicit representations and derive transductive generalization guarantees under suitable regularity conditions.

To investigate the statistical advantages of expectile aggregation, we establish two classification risk results under heterophilic neighborhood models. First, we consider a fixed neighborhood mixing profile and show that, under suitable conditions, an expectile-based nearest-center classifier achieves a lower classification risk than even the optimal classifier based on mean aggregation for sufficiently large neighborhood sizes. Second, motivated by the low label informativeness observed in several heterophilic benchmark datasets \citep{platonov2023characterizing}, we consider a local asymptotic regime in which the class information carried by each individual neighbor progressively decreases as the neighborhood size increases. Under an additional moment condition, we show that a common expectile level achieves a strictly smaller limiting classification risk than mean aggregation when both use nearest-center classification. Consequently, this advantage also holds for all sufficiently large finite neighborhood sizes. These results demonstrate how expectile aggregation can exploit class-dependent distributional information to improve classification, even when individual neighbors become weakly informative.

Finally, we evaluate ExpectileGCN on both homophilic and heterophilic benchmark datasets. In addition to standard node classification experiments, we examine its performance under feature corruption and structural perturbations. The experimental results demonstrate competitive predictive performance across different graph structures and improved robustness under the evaluated perturbation settings.

\subsection{Related works}

Research on neighborhood aggregation addresses several related objectives, including predictive expressiveness, adaptation to graph structure, and robustness to corrupted observations. Early GNN architectures commonly used mean or sum aggregation \citep{kipf2017semi,hamilton2017inductive}. To reduce sensitivity to graph perturbations, \citet{geisler2020reliable} proposed SoftMedoid aggregation, inspired by robust statistics, while \citet{chen2021median} studied median-based aggregation under structural perturbations. These approaches motivate the use of statistical aggregation operators to control how neighboring messages influence a node representation.

Another line of research combines multiple neighborhood statistics to capture different distributional characteristics. For example, \citet{corso2020principal} proposed Principal Neighbourhood Aggregation (PNA), which combines statistics such as the mean, maximum, minimum, and standard deviation with degree scalers. PNA learns to use these complementary summaries within the network. Our approach instead parameterizes the aggregation statistic itself through an expectile level, allowing its response to message deviations to vary continuously while preserving the output dimension of mean aggregation.

Parameterized aggregation also underlies GMGCN, which learns an $L_p$-based aggregation operator \citep{wang2023betagcn}. This approach illustrates how the aggregation behavior can be adapted through a trainable statistical parameter. Expectile aggregation uses a different parameterization: $\tau\in(0,1)$ controls asymmetric weighting around an implicitly determined location, recovers the mean at $\tau=\frac{1}{2}$, and shifts emphasis toward lower or upper messages on either side of this value. Its asymmetric least-squares definition applies directly to signed inputs. Our analysis connects this parameterization to classification risk through the scale and amplitude information present in neighborhood message distributions.

\subsection{Contributions}

We propose ExpectileGCN, a graph neural network with learnable expectile aggregation. The operator extends mean aggregation through asymmetric weighting of message deviations, accommodates positive and negative features directly, and retains one aggregated value per feature coordinate. For finite neighborhoods, varying the expectile level connects minimum-like, mean, and maximum-like aggregation behaviors, providing an interpretable parameter for adapting the statistical summary to the prediction task.

We develop a sigmoid-smoothed fixed-point formulation and an iterative implementation for end-to-end learning. For the implicit model, we establish existence, uniqueness, and parameter stability under suitable conditions. Theorems~\ref{thm:rademacher} and~\ref{thm:generalization} bound the transductive Rademacher complexity and the resulting training--test loss gap, respectively, linking the stability of the aggregation operator to generalization on a fixed graph.

We compare expectile and mean aggregation using identical input messages in two multiclass neighborhood models with self-inclusion. Theorem~\ref{thm:shape-risk} gives a finite-neighborhood risk-ratio bound under a fixed heterophilic mixing profile, yielding strict improvement over the optimal mean-based classifier when scale separation and neighborhood size satisfy the stated conditions. Theorem~\ref{thm:local-risk} considers mixing profiles approaching label independence as the neighborhood grows. Under a moment condition identifying additional amplitude information, a common expectile level achieves a strictly smaller limiting risk than mean aggregation when both use nearest-center classification. These results characterize how aggregation can exploit distributional information under fixed and diminishing per-neighbor class signals.

In summary, ExpectileGCN combines learnable distribution-sensitive aggregation with stability and generalization guarantees. Our classification risk analysis further demonstrates how the choice of aggregation statistic affects the extraction of neighborhood information under different heterophilic mixing regimes.\\
\textbf{Outline.} The remainder of this paper is organized as follows. Section~\ref{sec:model} introduces the ExpectileGCN framework, including expectile aggregation, its smooth approximation and fixed-point formulation, the two-layer network architecture, and computational complexity. Section~\ref{sec:theo} presents the theoretical analysis. We establish stability and transductive generalization guarantees for ExpectileGCN, and then study the classification risk of expectile aggregation from the perspective of neighborhood information, considering both informative and progressively weaker neighborhood signals. Section~\ref{sec:exper} reports experiments on homophilic and heterophilic benchmark datasets, including comparisons with baseline methods, robustness evaluations under feature and structural perturbations, and hyperparameter sensitivity analysis. Finally, Section~\ref{sec:con} concludes the paper and discusses directions for future research. Additional experimental results are provided in the appendix; all proofs and auxiliary lemmas are collected in the Supplementary Material.

\section{Estimation and Inference of ExpectileGCN}
\label{sec:model}
\subsection{Notation}
Let $\mathcal{G}=(\mathcal{V},\mathcal{E})$ denote a graph, where $\mathcal{V}$ and $\mathcal{E}$ are the sets of nodes and edges, respectively. We denote by $N=|\mathcal{V}|$ the total number of nodes in the graph. The graph structure is represented by an adjacency matrix $\mathbf{A}\in\{0,1\}^{N\times N}$, where $A_{ij}=1$ if there exists an edge between nodes $i$ and $j$, and $A_{ij}=0$ otherwise. For each node $i\in\mathcal{V}$, let $\mathcal N(i)=\{i\}\cup\{j:A_{ij}=1\}$ denote its aggregation neighborhood, including the node itself. Each node $i$ is associated with a feature vector $\mathbf{x}_i\in\mathbb{R}^q$, and the feature matrix of all nodes is denoted by $\mathbf{X}\in\mathbb{R}^{N\times q}$. We write $x_{j,m}$ for the $m$-th feature component of node $j$. In addition, each node is associated with a class label $y_i\in\mathcal{C}$, where $\mathcal{C}$ denotes the set of node classes and $|\mathcal{C}|=K$. For a given class $c\in\mathcal{C}$, let $\mathcal{V}_c=\{i\in\mathcal{V}:y_i=c\}$ denote the corresponding node subset. We consider the semi-supervised node classification setting, where labels are observed only for a subset of training nodes $\mathcal{V}_{\mathrm{train}}\subset\mathcal{V}$. Let $m=|\mathcal{V}_{\mathrm{train}}|$ denote the number of labeled training nodes. The goal is to learn a mapping $f:\mathcal{V}\rightarrow\mathcal{C}$, using the graph structure $\mathcal{G}$, node features $\mathbf{X}$, and the observed labels in $\mathcal{V}_{\mathrm{train}}$. Following standard graph neural network formulations, we define the row-normalized adjacency matrix with self-loops as $\tilde{\mathbf{A}}=\mathbf{D}^{-1}(\mathbf{A}+\mathbf{I})$,
where $\mathbf{I}$ is the identity matrix and $\mathbf{D}$ is the diagonal degree matrix of $\mathbf{A}+\mathbf{I}$. The entry $\tilde A_{ij}$ denotes the normalized propagation weight between nodes $i$ and $j$. Thus, $\tilde A_{ij}\ge0$ and $\sum_{j\in\mathcal N(i)}\tilde A_{ij}=1$ for every node $i$, including isolated nodes.
\subsection{Expectile Aggregation}
Building on the preceding notation, we now define the proposed expectile-based neighborhood aggregation operator. For a node $i\in\mathcal V$ and feature dimension $m\in\{1,\dots,q\}$, let $x_{j,m}$ denote the $m$-th feature component of a neighboring node $j\in\mathcal N(i)$, and let $\tilde A_{ij}$ denote the corresponding normalized adjacency weight. Given a fixed expectile level $\tau \in (0,1)$, we define the
aggregated feature $r^{\ast}_{i,m}$ as the minimiser of the asymmetric weighted
least-squares objective:
\begin{equation}
	r^{\ast}_{i,m}
	= \arg\min_{u \in \mathbb R}
	\sum_{j\in \mathcal N(i)}
	\tilde A_{ij}\,
	\bigl|\tau - \mathbf{1}\{x_{j,m} \le u\}\bigr|\,(x_{j,m} - u)^2.
	\label{eq:expectile-opt}
\end{equation}
For $\tau\in(0,1)$, the objective in Eq.~\eqref{eq:expectile-opt} is strictly convex and coercive because the normalized weights are nonnegative and sum to one. Hence the exact expectile exists uniquely, without a contraction assumption. Equivalently, $r_{i,m}$ can be characterized as the solution $r^{\ast}_{i,m} \in \mathbb R$ of the following implicit equation (obtained by setting the derivative of Eq.~\eqref{eq:expectile-opt} to zero):
\begin{equation}
	\sum_{j\in \mathcal N(i)}\tilde A_{ij}\alpha_{i,j}(r^{\ast}_{i,m})\, (x_{j,m}-r^{\ast}_{i,m})=0,
\end{equation}
where the asymmetric coefficient $\alpha_{i,j}(r^{\ast}_{i,m})$ is defined as
\begin{equation}
	\alpha_{i,j}(r^{\ast}_{i,m})
	=\begin{cases}
		\tau,       & x_{j,m} \ge r^{\ast}_{i,m},\\[2pt]
		1 - \tau,   & x_{j,m} < r^{\ast}_{i,m}.
		\label{eq:expectile-eq}
	\end{cases}
\end{equation}

\begin{figure}[H]
	\centering
	\IfFileExists{ExpectileGCN-Algorithm.pdf}{%
		\includegraphics[width=\textwidth,height=0.48\textheight,keepaspectratio]{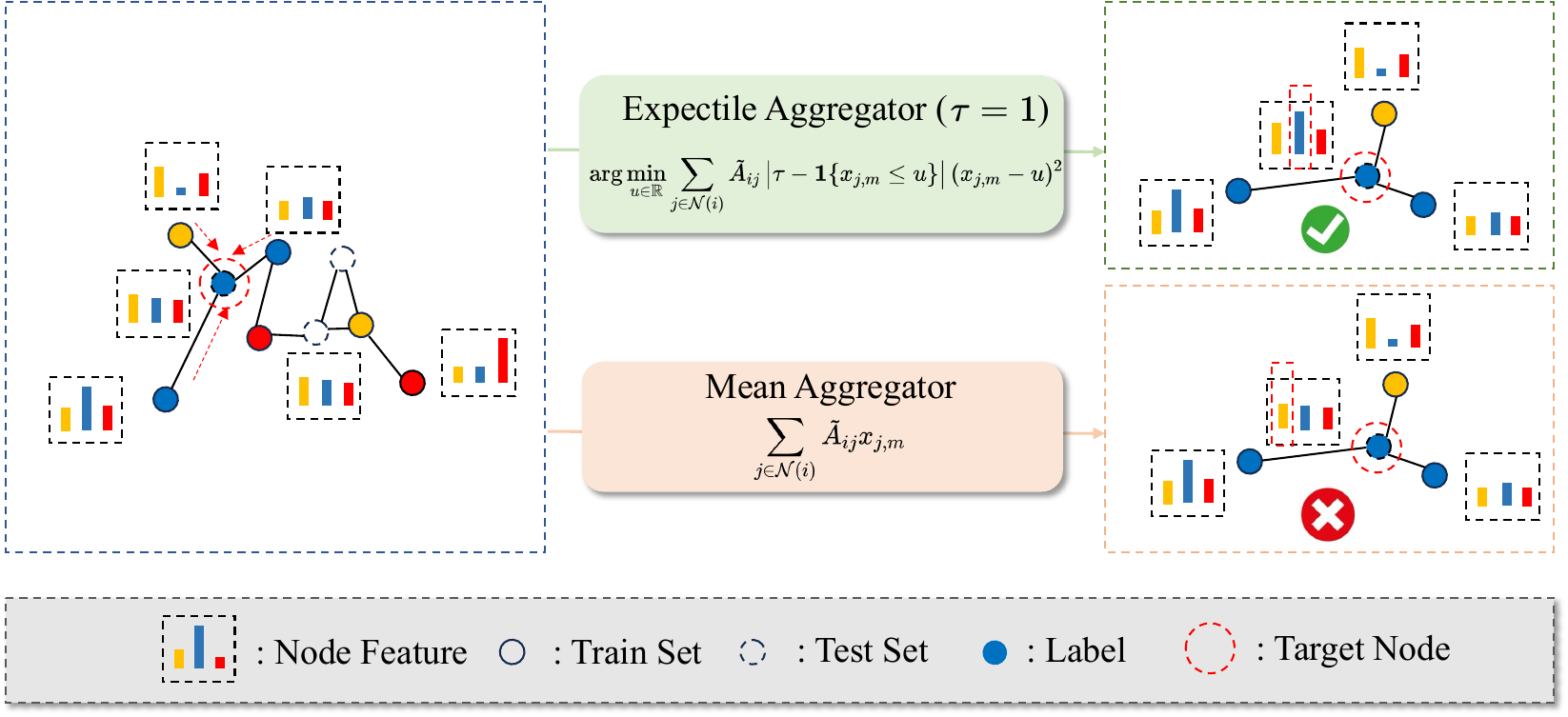}%
	}{
		\fbox{\parbox[c][0.48\textheight][c]{0.94\textwidth}{\centering
		Algorithm diagram reserved.\\[1em]
		The source asset \texttt{ExpectileGCN-Algorithm diagram.pdf}\\
		must be supplied before submission.}}%
	}
	\caption{~Illustration of the Expectile aggregation mechanism compared with conventional mean aggregation. Mean aggregation treats all neighboring signals uniformly, which may cause representation distortion under noisy or heterophilic neighborhoods. In contrast, ExpectileGCN adaptively emphasizes different parts of the neighborhood distribution through the expectile parameter $\tau$, enabling more flexible aggregation across different neighborhood distributions.}
	\label{fig:expectile_algorithm}
\end{figure}

Thus, the expectile serves as an adaptive, tail-sensitive neighbourhood summary: when $\tau = 0.5$, the aggregation reduces to the standard mean; when $\tau > 0.5$, the aggregation emphasises larger neighbour values; and when $\tau < 0.5$, smaller neighbour values receive greater emphasis. As illustrated in Figure~\ref{fig:expectile_algorithm}, this flexibility allows ExpectileGCN to move beyond uniform smoothing by selectively adjusting aggregation toward more informative structural patterns. The resulting aggregated representation of node $i$ in
dimension $m$ is therefore the value $r^{\ast}_{i,m}$ that balances neighbour residuals asymmetrically according to $\tau$. Collecting all feature dimensions yields the final expectile-aggregated
vector:
\[
\mathbf{r}^{\ast}_i = (r^{\ast}_{i,1},\dots,r^{\ast}_{i,q}) \in \mathbb R^q,
\]
which is the representation of node $i$ in this layer.

In addition, Figure~\ref{fig:expectile_algorithm} provides an intuitive illustration of the key motivation behind ExpectileGCN. Unlike the standard mean aggregator, which assigns equal influence to all neighboring signals and can therefore be highly sensitive to noisy, misleading, or structurally inconsistent neighbors, the expectile aggregator adaptively reweights neighbor contributions through the asymmetric coefficient controlled by $\tau$. This allows the model to adaptively emphasize different regions of the neighborhood message distribution, potentially capturing class-discriminative information that is not reflected in the mean.

\subsection{Numerical Approximation of Expectile Aggregation}

While the expectile-based aggregation defined in Eq.~\eqref{eq:expectile-opt} provides an adaptive neighborhood summary, its implicit definition requires a numerical procedure for integration into end-to-end neural network training. Specifically, the empirical expectile satisfies
\begin{equation}
\sum_{j\in\mathcal N(i)}
\tilde A_{ij}\,\alpha_{i,j}(r_{i,m})
(x_{j,m}-r_{i,m})=0.
\end{equation}
The asymmetric weights depend on the unknown aggregated value through indicator functions, resulting in a piecewise-defined fixed-point operator. We therefore develop a differentiable approximation based on sigmoid smoothing and fixed-point iterations, allowing the aggregation levels and network parameters to be optimized jointly.

We first replace the indicator function with a smooth sigmoid approximation:
\begin{equation}
\mathbf{1}\{x_{j,m}\ge r_{i,m}\}
\approx
\sigma\bigl(s(x_{j,m}-r_{i,m})\bigr),
\end{equation}
where $\sigma(x)=1/(1+e^{-x})$ is the sigmoid function and $s>0$ controls the approximation sharpness. As $s\to\infty$, the sigmoid converges to the indicator for $x_{j,m}\ne r_{i,m}$. At equality, the sigmoid equals $1/2$, but the corresponding residual vanishes and therefore does not affect the limiting expectile equation. For finite $s$, the smoothed asymmetric coefficient is
\begin{equation}
\begin{aligned}
\tilde\alpha_{i,j}(r_{i,m})=\tau\,\sigma\bigl(s(x_{j,m}-r_{i,m})\bigr)+
(1-\tau)
\left[1-\sigma\bigl(s(x_{j,m}-r_{i,m})\bigr)\right].
\end{aligned}
\label{eq:sigmod smooth}
\end{equation}
The resulting coefficient is continuously differentiable with respect to the aggregated value and input messages.

Using the smoothed coefficients, we compute the aggregated representation through fixed-point iterations. Starting from an initial value $r_{i,m}^{(0)}$, such as the normalized weighted mean, we iteratively calculate
\begin{equation}
\begin{aligned}
\tilde\alpha_{i,j}^{(t)}=
\tau\,\sigma\bigl(s(x_{j,m}-r_{i,m}^{(t)})\bigr)+
(1-\tau)
\left[1-\sigma\bigl(s(x_{j,m}-r_{i,m}^{(t)})\bigr)\right],
\end{aligned}
\end{equation}
followed by
\begin{equation}
r_{i,m}^{(t+1)}
=
\frac{
\sum_{j\in\mathcal N(i)}
\tilde A_{ij}\tilde\alpha_{i,j}^{(t)}x_{j,m}
}{
\sum_{j\in\mathcal N(i)}
\tilde A_{ij}\tilde\alpha_{i,j}^{(t)}
}.
\end{equation}
Each iteration updates the aggregated representation using asymmetric weights determined by the current estimate. After $T$ iterations, $r_{i,m}^{(T)}$ serves as the numerical output of the aggregation layer.

This procedure implements expectile-based aggregation through differentiable weighted averaging operations. The sharpness parameter $s$ controls the approximation of the original asymmetric weighting rule, while the iteration count $T$ determines the number of fixed-point updates. Under the contraction conditions established in the theoretical analysis, the iteration converges to a unique smoothed fixed point. The resulting approximation can be incorporated into standard GNN architectures and optimized jointly with the network parameters.

\subsection{ExpectileGCN}
Building upon the expectile aggregation operator introduced in the previous subsection, we now describe the overall GNN architecture used in this work. Since existing robust aggregation frameworks such as \citet{geisler2020reliable}, \citet{chen2021median}, and \citet{wang2023betagcn} are primarily developed and evaluated under two-layer graph neural network architectures, we also adopt a two-layer ExpectileGCN formulation for consistency and comparability. The subsequent theoretical analysis and experimental evaluation throughout this paper follow the same two-layer setting. Let the input feature matrix be $\mathbf{X}\in\mathbb{R}^{N\times q}$, where each node $i$ is associated with a feature vector $\mathbf{x}_i\in\mathbb{R}^{q}$. The first and second layers use learnable weight matrices $\mathbf{W}^{(1)}\in\mathbb{R}^{q\times q_1}$ and $\mathbf{W}^{(2)}\in\mathbb{R}^{q_1\times K}$, respectively. The expectile parameters for the two layers are denoted by $\tau^{(1)}\in(0,1)$ and $\tau^{(2)}\in(0,1)$, respectively.

\paragraph{First-layer expectile aggregation.}
For the first layer, we write the expectile aggregation operator in component form.
Let $\mathbf{z}_j^{(1)}=(z^{(1)}_{j,1},\dots,z^{(1)}_{j,q_1})^\top\in\mathbb{R}^{q_1}$ be the
linearly transformed message of node $j$, where
$\mathbf{z}_j^{(1)}=\left(\mathbf{x}_j^\top\mathbf{W}^{(1)}\right)^\top$ and
$\mathbf{W}^{(1)}=(\mathbf{w}^{(1)}_{1},\cdots,\mathbf{w}^{(1)}_{q_1})\in\mathbb{R}^{q\times q_1}$. Define the operator $T_{\tau^{(1)},\mathbf{W}^{(1)}}(\cdot;\mathbf{X},\tilde{\mathbf{A}}):\mathbb{R}^{q_1}\to\mathbb{R}^{q_1}$ componentwise as
\begin{equation}
\begin{aligned}
&T_{\tau^{(1)},\mathbf{W}^{(1)}}(\mathbf{r}_i^{(1)};\mathbf{X},\tilde{\mathbf{A}})=
\left(
T_{\tau^{(1)},\mathbf{w}_{1}^{(1)}}(r_{i,1}^{(1)};\mathbf{X},\tilde{\mathbf{A}}),
\ldots,
T_{\tau^{(1)},\mathbf{w}_{q_1}^{(1)}}(r_{i,q_1}^{(1)};\mathbf{X},\tilde{\mathbf{A}})
\right)^\top.
\end{aligned}
\end{equation}

Each component $T_{\tau^{(1)},\mathbf{w}^{(1)}_{m}}(r^{(1)}_{i,m};\mathbf{X},\tilde{\mathbf{A}})$,
corresponding to feature coordinate $m\in\{1,\dots,q_1\}$, is given by
\[
\frac{\displaystyle
	\sum_{j\in\mathcal{N}(i)}
	\tilde A_{ij}\,
	\tilde{\alpha}(r_{i,m}^{(1)},\tau^{(1)},\mathbf{w}_m^{(1)};\mathbf{x}_j)\,
	z^{(1)}_{j,m}
}
{\displaystyle
	\sum_{j\in\mathcal{N}(i)}
	\tilde A_{ij}\,
	\tilde{\alpha}(r_{i,m}^{(1)},\tau^{(1)},\mathbf{w}_m^{(1)};\mathbf{x}_j)
},
\]
where the smoothed asymmetric weight for neighbor $j$ and coordinate $m$ is
defined elementwise by
\begin{equation}
	\begin{aligned}
			\tilde{\alpha}(r_{i,m}^{(1)},\tau^{(1)},\mathbf{w}_m^{(1)};\mathbf{x}_j)=\tau^{(1)}\,\sigma\!\big(s(z^{(1)}_{j,m}-r_{i,m}^{(1)})\big)+(1-\tau^{(1)})\left[1-\sigma\!\big(s(z^{(1)}_{j,m}-r_{i,m}^{(1)})\big)\right].
		\nonumber
	\end{aligned}
\end{equation}
Here $\mathbf{r}_{i}^{(1)}=(r^{(1)}_{i,1},\dots,r^{(1)}_{i,q_1})^\top\in\mathbb{R}^{q_1}$ denotes the candidate fixed point for node $i$ at layer one, and $\sigma$
denotes the sigmoid applied componentwise. By construction, the vector
operator is coordinate separable: each coordinate $m$ uses only
$z^{(1)}_{j,m}$ and $r_{i,m}$. The first layer aggregated representation of node $i$ is the fixed point of this vector operator. Denoting the first layer fixed point by
$\mathbf{r}_i^{(1)\ast}=(r_{i,1}^{(1)\ast},\cdots,r_{i,q_1}^{(1)\ast})^\top\in\mathbb{R}^{q_1}$, we have $\mathbf{r}_i^{(1)\ast}=T_{\tau^{(1)},\mathbf{W}^{(1)}}\!\big(\mathbf{r}_i^{(1)\ast};\mathbf{X}\big)$. Collecting the fixed-point representations of all nodes, we obtain the layer-1 aggregated feature matrix $\mathbf{R}^{(1)\ast}=\big(\mathbf{r}_1^{(1)\ast},\ldots,\mathbf{r}_N^{(1)\ast}\big)^\top
\in\mathbb{R}^{N\times q_1}$. Let $\phi(\cdot)$ denote a pointwise activation function. In this work, we choose $\phi(\cdot)=\mathrm{ReLU}(\cdot)$.

\paragraph{Second-layer expectile aggregation.}
The second layer applies the same expectile aggregation mechanism as the first
layer, but operates on the hidden representations obtained after nonlinear
activation. Specifically, each node $j$ is associated with the activated first-layer
representation $\phi(\mathbf r_j^{(1)\ast})\in\mathbb R^{q_1}$.
Given a weight matrix $\mathbf W^{(2)}=(\mathbf w^{(2)}_1,\dots,\mathbf w^{(2)}_K)\in\mathbb R^{q_1\times K}$, the transformed message is the column vector $\mathbf z_j^{(2)}=\left(\phi(\mathbf r_j^{(1)\ast})^\top\mathbf W^{(2)}\right)^\top\in\mathbb R^K$. Its coordinate $m$ is the scalar $z^{(2)}_{j,m} = \phi(\mathbf r_j^{(1)\ast})^\top \mathbf w^{(2)}_m$. The second-layer aggregation operator
$T_{\tau^{(2)},\mathbf W^{(2)}}(\cdot;\phi(\mathbf R^{(1)\ast}),\tilde{\mathbf A})
:\mathbb R^K\to\mathbb R^K$ is defined componentwise using the same smoothed expectile weights as in the first layer. As before, the operator is coordinate separable and admits a fixed-point characterization. Denoting the fixed point by $\mathbf r_i^{(2)\ast}=(r^{(2)\ast}_{i,1},\dots,r^{(2)\ast}_{i,K})^\top\in\mathbb R^K$, we obtain the final node-level output representations. Collecting all nodes yields the second-layer output matrix
$\mathbf R^{(2)\ast}\in\mathbb R^{N\times K}$, whose rows are used as logits for
node classification.

The second layer aggregated output for node $i$ is
$\mathbf{r}_i^{(2)\ast}=T_{\tau^{(2)},\mathbf{W}^{(2)}}(\mathbf{r}_i^{(2)\ast};\phi\left(\mathbf{R}^{(1)\ast}\right),\tilde{\mathbf{A}})$.
Collecting the fixed-point outputs of all nodes from the second layer, we obtain the final representation matrix $\mathbf{R}^{(2)\ast}=\big(\mathbf{r}_1^{(2)\ast},\ldots,\mathbf{r}_N^{(2)\ast}\big)^\top\in\mathbb{R}^{N\times K}$. Each vector $\mathbf{r}_i^{(2)\ast}$ serves as the logit vector for node $i$ in a $K$-class classification task.
The predicted class probability is obtained via the softmax function, $\hat{\mathbf{y}}_i=\mathrm{softmax}\!\left(\mathbf{r}_i^{(2)\ast}\right)$. Let $\mathbf{y}_i\in\{0,1\}^{K}$ denote the one-hot encoded ground-truth label of
node $i$. We adopt the standard categorical cross-entropy loss $\ell\!\left(\hat{\mathbf{y}}_i,\mathbf{y}_i\right)=- \sum_{k=1}^{K} y_{i,k}\ln \hat{y}_{i,k}$. For the theoretical analysis, we write the same loss as a function of the logits, $g(\mathbf z,\mathbf y):=\ell(\mathrm{softmax}(\mathbf z),\mathbf y)=-\mathbf y^\top\mathbf z+\log\!\left(\sum_{k=1}^{K}e^{z_k}\right)$, where $\mathbf z\in\mathbb R^K$ and $\mathbf y$ is a one-hot label.
Accordingly, the loss function is defined as
\begin{equation}
	\begin{aligned}
		\label{eq:loss}
		\widehat{\mathcal{L}}_m(\theta)=\frac{1}{m}\sum_{i\in\mathcal V_{\mathrm{train}}}g\!\left(\mathbf{r}_i^{(2)\ast}(\theta),\mathbf{y}_i\right),
	\end{aligned}
\end{equation}
where $\theta$ denotes the collection of all model parameters across layers. Recall that $\mathcal V_{\mathrm{train}}\subset\mathcal V$ denotes the set of labeled training nodes with cardinality $|\mathcal V_{\mathrm{train}}|=m$. The two-layer ExpectileGCN model computes node representations through graph message passing and produces a second-layer fixed-point output $\mathbf R^{(2)\ast}=(\mathbf r_1^{(2)\ast},\ldots,\mathbf r_N^{(2)\ast})^\top \in\mathbb R^{N\times K}$.

\subsection{Computational Complexity.}
We compare the per-layer aggregation complexity of expectile, median, and SoftMedoid aggregation. Let $N$ and $|\mathcal E|$ denote the numbers of nodes and edges, respectively, $d_{\max}$ the maximum node degree excluding self-loops, and $T$ the number of fixed-point iterations. We focus on aggregation costs, excluding linear transformations and feature dimensions.

Expectile aggregation updates the asymmetric weights and aggregates messages along each edge at every fixed-point iteration. Each iteration therefore requires $O(N+|\mathcal E|)$ operations, including self-messages, giving a total complexity of $O(T(N+|\mathcal E|))$. For fixed $T$, the aggregation cost is linear in graph size, with $O(N+|\mathcal E|)$ memory for sparse graph storage and node representations. For coordinate-wise median aggregation, a sorting-based implementation requires $O\left(\sum_{v=1}^{N}(d_v+1)\log(d_v+1)\right)=O((N+|\mathcal E|)\log(d_{\max}+1))$ time per layer, where $d_v$ is the degree of node $v$ excluding self-loops and $d_{\max}=\max_v d_v$. SoftMedoid aggregation instead evaluates pairwise distances among messages within each neighborhood, requiring $O\left(\sum_{v=1}^{N}(d_v+1)^2\right)=O(N(d_{\max}+1)^2)$ time in the worst case. Explicitly storing these pairwise distances can incur comparable memory costs. The neighbor truncation strategy of \citet{geisler2020reliable} reduces this cost to $O(Nk^2)$ by retaining $k$ messages per neighborhood. Unlike SoftMedoid, expectile aggregation requires no pairwise neighborhood comparisons. Its linear dependence on graph size for fixed $T$ makes it particularly suitable for large sparse graphs and neighborhoods with high degree.

\section{Theoretical Analysis of ExpectileGCN}
\label{sec:theo}
\subsection{Definitions and Assumptions}
\label{definitions}
In this section, we develop an upper bound on the transductive Rademacher complexity and the resulting same-graph generalization gap for the proposed two-layer ExpectileGCN by characterizing the hypothesis class induced by its implicit aggregation mechanism. To facilitate the subsequent complexity and generalization analysis, we first formalize the class of logit functions generated by ExpectileGCN under the parameter space $\Theta$. Throughout the smoothed-model analysis, fix an arbitrary sharpness parameter $s>0$ in Eq.~\eqref{eq:sigmod smooth}, common to both layers and fixed before the random partition. It is not part of the learned parameter vector $\theta$; its dependence is suppressed in the notation for the operators, fixed points, and hypothesis class. Specifically, we denote by $\mathcal H$ the hypothesis class induced by ExpectileGCN before the softmax transformation, defined as
\begin{equation}
	\begin{aligned}
		\mathcal{H} = \Bigl\{\, h_{\theta} : (\mathbf X,\tilde{\mathbf A},i)\mapsto \mathbf r_i^{(2)\ast}(\theta),\ i\in\mathcal V,\ \theta=(\theta^{(1)},\theta^{(2)})\in\Theta \,\Bigr\}.
		\nonumber
	\end{aligned}
\end{equation}
We analyze a fixed graph, fixed feature matrix, and fixed labels on a node set $\mathcal V_0\subseteq\mathcal V$ of size $m+u$. The only randomness in the generalization statement is a uniformly random partition without replacement of $\mathcal V_0$ into a training set $\mathcal V_{\mathrm{train}}$ of size $m$ and a test set $\mathcal V_{\mathrm{test}}$ of size $u$, following the transductive setting of \citet{tang2023towards}. The parameter set $\Theta$, the induced hypothesis class $\mathcal H$, and all uniform bounds are fixed before the random partition is drawn; the selected parameter $\theta$ may depend on the observed training labels. Graph message passing may use the entire fixed graph; the split only determines which labels enter the training and test risks. For notational convenience, let $\mathbf R^{(2)\ast}_{0}(\theta):=\left(\mathbf r_i^{(2)\ast}(\theta)\right)_{i\in\mathcal V_0}\in\mathbb R^{(m+u)\times K}$ and $\mathcal M_0=\left\{\mathbf R^{(2)\ast}_{0}(\theta)\;\middle|\;\theta\in\Theta\right\}$. We use $g\circ\mathcal H$ to denote the resulting scalar loss class. Define
\begin{equation}
	\widehat{\mathcal L}_m(\theta)=\frac{1}{m}\sum_{i\in\mathcal V_{\mathrm{train}}}g\!\left(\mathbf r_i^{(2)\ast}(\theta),\mathbf y_i\right),\qquad
	\widehat{\mathcal L}_u(\theta)=\frac{1}{u}\sum_{i\in\mathcal V_{\mathrm{test}}}g\!\left(\mathbf r_i^{(2)\ast}(\theta),\mathbf y_i\right).
	\nonumber
\end{equation}

\begin{definition}\label{de:coveringnb}
	Let $(\mathcal{M}, d)$ be a metric space and $r>0$.
	The covering number $\mathcal{N}(\mathcal{M}, d, r)$ is defined as the smallest integer $n \in \mathbb{N}$ such that there exist points $m_1,\dots,m_n \in \mathcal{M}$ satisfying $\mathcal{M} \subset \bigcup_{i=1}^{n} B_d(m_i, r)$, where $B_d(m_i, r)=\{m\in\mathcal{M}: d(m,m_i)\le r\}$ denotes the closed ball of radius $r$ centered at $m_i$. When the metric $d$ is induced by a norm $\|\cdot\|$, we write $\mathcal{N}(\mathcal{M},\|\cdot\|,r)$.
\end{definition}

\begin{definition}\label{de:em_rademacher}
	Let $p_0=\frac{mu}{(m+u)^2}$. For a class of scalar functions $\mathcal F$ evaluated on the $m+u$ fixed node examples $z_1,\ldots,z_{m+u}$, its transductive Rademacher complexity is
	\begin{equation}
		\begin{aligned}
			\mathfrak R_{m+u}^{\mathrm{tr}}(\mathcal F)
			=\left(\frac1m+\frac1u\right)
			\mathbb E_{\boldsymbol\sigma}\!\left[\sup_{f\in\mathcal F}\sum_{i=1}^{m+u}\sigma_i f(z_i)\right],
		\end{aligned}
	\end{equation}
	where the coordinates of $\boldsymbol\sigma$ are independent and satisfy $\Pr(\sigma_i=1)=\Pr(\sigma_i=-1)=p_0$ and $\Pr(\sigma_i=0)=1-2p_0$.
\end{definition}

For ExpectileGCN, Definition~\ref{de:em_rademacher} gives
\begin{equation}
	\begin{aligned}
		\mathfrak R_{m+u}^{\mathrm{tr}}(g\circ\mathcal H)
		=\left(\frac1m+\frac1u\right)
		\mathbb E_{\boldsymbol\sigma}\!\left[\sup_{\theta\in\Theta}\sum_{i=1}^{m+u}\sigma_i\,g\!\left(\mathbf r_i^{(2)\ast}(\theta),\mathbf y_i\right)\right].
		\nonumber
	\end{aligned}
\end{equation}
For any $s>0$, the smoothed indicator satisfies $|\partial_z\sigma(s(z-r))|=|\partial_r\sigma(s(z-r))|\le \frac{s}{4}$. This derivative bound determines the dependence of the operator Lipschitz constants on $s$. The following assumptions provide uniform bounds over $\theta\in\Theta$ for each such $s$.
\begin{assumption}
	\label{assump:x_bound}
	The $\ell_2$-norm of the feature vector in the input space is bounded, namely, for some constant $C_x>0$, we assume $\left \| \mathbf{x}_i \right \|\le C_x$ for all nodes $i$.
\end{assumption}

\begin{assumption}
	\label{assump:parameter-space}
	For the parameters of two layers $\theta^{(1)} = (\operatorname{vec}\left(\mathbf{W}^{(1)}\right)^\top, \tau^{(1)})^\top$ and $\theta^{(2)} = (\operatorname{vec}\left(\mathbf{W}^{(2)}\right)^\top, \tau^{(2)})^\top$, we assume the following conditions hold:\\
	(a) Each coordinate weight vector is bounded in $\ell_2$-norm, namely $\|\mathbf{w}^{(1)}_m\| \le C_w$ and $\|\mathbf{w}_k^{(2)}\| \le C_w$ for all $1 \le m \le q_1$ and $1\le k\le K$, where $C_w>0$ is a constant.\\
	(b) The expectile level of each layer is bounded away from the degenerate values $0$ and $1$: $\epsilon_{\tau} \le \tau^{(l)} \le 1-\epsilon_{\tau}$ for some constant $0< \epsilon_{\tau} \le \frac{1}{2}$.
\end{assumption}

\begin{assumption}
	\label{assump:graph}
	The underlying graph $\mathcal{G}=(\mathcal{V},\mathcal{E})$ is a simple undirected graph without self-loops or multi-edges. Self-loops are added only for aggregation through $\mathbf A+\mathbf I$. The aggregation neighborhoods have uniformly bounded size: $1\le|\mathcal N(i)|=\deg_{\mathcal G}(i)+1\le d$ for all $i\in\mathcal V$, where $d\ge1$ includes the self-node.
\end{assumption}

\begin{assumption}[Uniform contraction margin]
	\label{assump:contraction}
	For the fixed sharpness $s>0$, there exists a constant $\kappa=\kappa(s)\in(0,1)$, independent of the node, coordinate, and parameter value, such that for every $\theta\in\Theta$ both layer-wise scalar aggregation maps are contractions with modulus at most $1-\kappa$. Specifically, for all admissible $r,r'\in\mathbb R$,
	\begin{equation}
		\begin{aligned}
		\left|T_{\tau^{(1)},\mathbf w_m^{(1)}}(r;\mathbf X,\tilde{\mathbf A})-T_{\tau^{(1)},\mathbf w_m^{(1)}}(r';\mathbf X,\tilde{\mathbf A})\right|&\le(1-\kappa)|r-r'|,\\
		\left|T_{\tau^{(2)},\mathbf w_k^{(2)}}(r;\phi(\mathbf R^{(1)\ast}),\tilde{\mathbf A})-T_{\tau^{(2)},\mathbf w_k^{(2)}}(r';\phi(\mathbf R^{(1)\ast}),\tilde{\mathbf A})\right|&\le(1-\kappa)|r-r'|,
		\end{aligned}
		\nonumber
	\end{equation}
	for every admissible first-layer fixed-point matrix $\mathbf R^{(1)\ast}$. The contraction margin $\kappa(s)$ is uniform over $\Theta$ for each $s$.
\end{assumption}

Assumptions~\ref{assump:x_bound} and~\ref{assump:parameter-space} impose standard boundedness conditions on the input features and model parameters, which are commonly used in generalization analyses of neural networks and graph neural networks to control hypothesis complexity \citep{liao2020pac,lv2021generalization,Zhou2025magnn,fung2026generalization}. These conditions also imply uniform boundedness of the implicit representations, as stated in the following proposition.

\begin{boundednessproposition}[Uniform boundedness of the implicit representations]
	\label{prop:r-bound}
	Under Assumptions~\ref{assump:x_bound} and~\ref{assump:parameter-space}, every first- and second-layer fixed point satisfies
	\[
	\left\|\mathbf r_i^{(1)\ast}\right\|
	\leq \sqrt{q_1}\,C_xC_w,
	\qquad
	\left\|\mathbf r_i^{(2)\ast}\right\|
	\leq \sqrt{Kq_1}\,C_xC_w^2.
	\]
	Consequently, we define $C_r:=\max\left\{\sqrt{q_1}\,C_xC_w,
	\sqrt{Kq_1}\,C_xC_w^2\right\}$, so that both fixed-point representations are uniformly bounded by $C_r$ for all nodes and all $\theta\in\Theta$.
\end{boundednessproposition}

Proposition~\ref{prop:r-bound} provides the uniform representation bound used throughout the subsequent complexity analysis. Its proof is provided in Section~S1 of the Supplementary Material. Assumption~\ref{assump:graph} additionally controls the aggregation neighborhood size, while Assumption~\ref{assump:contraction} ensures uniform stability of the implicit aggregation operators. In particular, the contraction margin $\kappa$ guarantees unique fixed points and controls their sensitivity to model parameters, which is required for the covering-number argument below. Using Proposition~\ref{prop:r-bound}, an explicit sufficient condition for Assumption~\ref{assump:contraction}, when $0<\epsilon_\tau<\frac{1}{2}$, is
\[
\max(C_x,C_r)\,C_w
\leq(1-\kappa)\frac{4\epsilon_\tau^2}{s(1-2\epsilon_\tau)(1+\epsilon_\tau)}.
\]
For $\epsilon_\tau=\frac{1}{2}$, both operators have modulus zero and no additional parameter restriction is required. Under Assumption~\ref{assump:contraction}, the Banach fixed-point theorem guarantees a unique solution to each layer-wise scalar equation and convergence of the corresponding fixed-point iteration. The fixed-point arguments and the verification of the sufficient condition are provided in Sections~S1 and~S4 of the Supplementary Material. The condition becomes more stringent as the fixed sharpness $s$ increases.

\subsection{Generalization Analysis}

We next study how the architectural complexity and the expectile mechanism of ExpectileGCN affect its generalization behavior. Since node classification is performed on a fixed graph and the training and test nodes are obtained by partitioning the same set of nodes, we adopt a transductive generalization framework, following the setting commonly used in theoretical analyses of GNNs \citep{tang2023towards}. Our analysis first controls the capacity of the induced hypothesis class through its transductive Rademacher complexity and then translates this complexity control into a bound on the discrepancy between training and test losses.

\setcounter{theorem}{0}
\begin{theorem}
	\label{thm:rademacher}
	Fix $s>0$ and let $L_{r^{(2)},\theta}$ be the uniform logit Lipschitz constant from Lemma~S13 of the Supplementary Material, evaluated at this $s$ and its contraction margin $\kappa(s)$. Define $p=q_1(q+K)+2$ and $C_\mathrm{params}=\sqrt{C_w^2(q_1+K)+2(1-\epsilon_{\tau})^2}$. Under Assumptions~\ref{assump:x_bound}--\ref{assump:contraction}, we have
	\begin{equation}
		\mathfrak R_{m+u}^{\mathrm{tr}}(g\circ\mathcal H)
		\le 8C_r\left(\frac1m+\frac1u\right)\sqrt{(m+u)p}
		\sqrt{1+\ln\left(1+\frac{4L_{r^{(2)},\theta}C_\mathrm{params}}{C_r}\right)} .
		\nonumber
	\end{equation}
\end{theorem}

Theorem~\ref{thm:rademacher} characterizes how the architecture, expectile levels, and stability of ExpectileGCN affect its hypothesis class complexity. The effective parameter dimension $p=q_1(q+K)+2$ captures the contribution of network width and input and output dimensions, while $C_{\mathrm{params}}$ incorporates the parameter norm bounds and the constraint $\epsilon_\tau\leq\tau\leq1-\epsilon_\tau$. The smoothing sharpness $s$ and contraction margin $\kappa(s)$ enter the bound through the parameter Lipschitz constant $L_{r^{(2)},\theta}$, which controls the sensitivity of the logits to changes in model parameters. For fixed architectural and stability constants, when the training and test sets occupy nonvanishing fractions of the graph, the transductive Rademacher complexity decreases at the rate $\mathcal{O}((m+u)^{-1/2})$.

The resulting complexity control directly yields a bound on the discrepancy between training and test performance under random sampling without replacement.

\begin{theorem}
	\label{thm:generalization}
	For the same fixed $s>0$ and $L_{r^{(2)},\theta}$ as in Theorem~\ref{thm:rademacher}, let $S_{m,u}=\frac{2\max\{m,u\}(m+u)}{\left(m+u-\frac{1}{2}\right)\left(2\max\{m,u\}-1\right)}$ and $c_0=\sqrt{\frac{32\ln(4e)}{3}}$. Under Assumptions~\ref{assump:x_bound}--\ref{assump:contraction}, with probability at least $1-\delta$ over the uniformly random partition without replacement, the following holds simultaneously for every $\theta\in\Theta$:
	\begin{equation}
		\begin{aligned}
			\widehat{\mathcal L}_u(\theta)
			\le{}&\widehat{\mathcal L}_m(\theta)
			+8C_r\left(\frac1m+\frac1u\right)\sqrt{(m+u)p}
			\sqrt{1+\ln\left(1+\frac{4L_{r^{(2)},\theta}C_\mathrm{params}}{C_r}\right)}
			\\
			&\quad+C_\ell c_0\left(\frac1m+\frac1u\right)\sqrt{\min\{m,u\}}
			+C_\ell\sqrt{\frac{S_{m,u}}{2}\left(\frac1m+\frac1u\right)\ln\frac{2}{\delta}},
			\nonumber
		\end{aligned}
	\end{equation}
	where $C_\ell=\log\!\bigl(1+(K-1)e^{2C_r}\bigr)$ uniformly bounds each loss.
\end{theorem}

Theorem~\ref{thm:generalization} establishes a uniform transductive generalization guarantee for ExpectileGCN, showing that its test loss is bounded by the training loss and three finite-sample terms. The first term measures model complexity through the transductive Rademacher bound in Theorem~\ref{thm:rademacher}. It explicitly accounts for the network dimension $p$, the bounded parameter space $C_\mathrm{params}$, and the parameter sensitivity $L_{r^{(2)},\theta}$ of the implicit representations. In particular, the contraction condition controls this sensitivity, linking the stability of expectile aggregation directly to the generalization bound. The second term captures the finite-sample correction associated with the transductive training and test split, while the third quantifies the sampling uncertainty at confidence level $1-\delta$. When the training and test sets occupy fixed positive proportions of the graph, all three terms converge to zero at the rate $\mathcal O\!\left((m+u)^{-\frac{1}{2}}\right)$, provided the model constants remain bounded. Thus, the difference between training and test losses becomes smaller as the number of nodes increases. Importantly, this guarantee applies uniformly over the admissible expectile levels and network parameters, providing theoretical support for learning the aggregation behavior. The result also highlights the role of contraction stability in controlling model complexity and controlling the transductive generalization gap of the smoothed ExpectileGCN hypothesis class. The analysis follows the transductive node classification setting, where training and test nodes are sampled without replacement from a fixed graph, as commonly considered in GNN generalization theory \citep{tang2023towards}. The proofs of Theorems~\ref{thm:rademacher} and~\ref{thm:generalization} are provided in Section~S2 of the Supplementary Material.

\subsection{Neighborhood Informativeness and Aggregation Risk}
\label{sec:aggregation-risk}

The effectiveness of neighborhood aggregation depends not only on the information contained in neighboring nodes, but also on how that information is summarized. Recent studies show that heterophilic neighborhoods can remain informative when their class distributions are distinguishable \citep{ma2022homophily,platonov2023characterizing}. However, conventional mean aggregation captures only the average neighborhood message and may overlook class information expressed through other distributional characteristics. This motivates a theoretical investigation of whether expectile aggregation can extract neighborhood information more effectively than mean aggregation and thereby reduce classification risk.

We investigate this issue through two complementary classification risk comparisons. Theorem~\ref{thm:shape-risk} considers neighborhoods with a fixed informative structure and shows that expectile aggregation can exploit differences in message scale to achieve a lower classification risk than mean aggregation. Theorem~\ref{thm:local-risk} considers a more challenging setting in which the information contributed by each neighbor progressively weakens and establishes that an expectile advantage can persist even in this regime. Throughout this subsection, we consider $K\geq3$ equally likely classes. Conditional on the root label $Y=y$ and aggregation size $D=d$, the neighboring messages $Z_1,\ldots,Z_{d-1}$ are independent draws from
\begin{equation}
F_y=\sum_{b=1}^K H_{yb}G_b,
\qquad G_b=\mathcal N(\mu_b,\sigma_b^2),
\end{equation}
where $H_{yb}$ is the probability of observing a class-$b$ neighbor. The self-message $Z_0$ is independently drawn from $G_y$, and the distribution of $D$ is common across classes. All $d$ messages receive equal weight $\frac{1}{d}$, consistent with the aggregation rule based on $\mathbf A+\mathbf I$.

We compare the empirical expectile $\widehat r_{\tau,d}$ with the mean $\overline Z_d=\widehat r_{\frac{1}{2},d}$ using the same neighborhood messages. Their population centers are determined by $\widetilde F_{y,d}=\frac{G_y+(d-1)F_y}{d}$. Here $d=|\mathcal N(i)|$ includes the self-node, as in Section~\ref{definitions}. This setting isolates the effect of the aggregation statistic: both methods observe identical messages, but summarize their distributions differently.

\subsubsection{Fixed Neighborhood Information: A Finite Neighborhood Risk Bound}

We first consider neighborhoods whose class information remains fixed as the neighborhood size increases. Let $\tau\in(\frac{1}{2},1)$ and consider the mixing profile $H_{yy}=\eta$ and $H_{yb}=\frac{1-\eta}{K-1}$ for $b\neq y$. When $\eta<\frac{1}{K}$, the graph is heterophilic, but its neighborhood distributions remain informative because the mixing probabilities differ across root classes.

Our main interest is a setting in which the class means are similar, but the message scales differ. In this case, mean aggregation may obscure class differences, whereas expectile aggregation can capture them through asymmetric weighting. Define $\varepsilon=\max_b|\mu_b-\mu_0|$ for some common location $\mu_0$, $\delta_\sigma=\min_{b<K}(\sigma_{b+1}-\sigma_b)$, $\rho=\frac{\sigma_K}{\sigma_1}$, and $q_H=\frac{1-K\eta}{K-1}$. After incorporating the self-message, the effective contrast between off-diagonal and diagonal mixture weights becomes $q_{H,d}=\frac{(d-1)q_H-1}{d}$. For small neighborhoods, the self-message can offset or even reverse the heterophilic mixing pattern, resulting in an effectively homophilic aggregation distribution. We therefore consider $d\geq d_0$, where $d_0\geq2$ is a fixed threshold satisfying $(d_0-1)q_H>1$, so that the heterophilic contribution of neighboring nodes dominates the self-message. This restriction allows us to focus on how expectile aggregation exploits information from heterophilic neighborhoods.

\begin{assumption}[Scale separation with a self-message]
\label{assump:risk-fixed}
The neighbor profile is heterophilic, $0\leq\eta<\frac{1}{K}$, and the class scales satisfy $0<\sigma_1<\cdots<\sigma_K$. The minimum aggregation size satisfies $(d_0-1)q_H>1$. Moreover, $\varepsilon<
\min\left\{\frac{\sigma_1}{2}, c(\tau,\rho)q_{H,d_0}\delta_\sigma\right\}$, where
$c(\tau,\rho)=\frac{(1-\tau)(2\tau-1)\varphi(\rho t_\tau)}
{2\sqrt5\,\tau^2\rho}>0$,
$\varphi$ is the standard normal density, and
$t_\tau=r_\tau(\mathcal N(0,1))>0$.
\end{assumption}

Assumption~\ref{assump:risk-fixed} ensures that differences in neighborhood scale dominate differences in location. When all component means coincide, the mean aggregates have identical population centers across classes, while expectile aggregation produces distinct centers by responding to scale differences. The assumption allows this advantage to persist when the component means differ slightly. 

For Theorem~\ref{thm:shape-risk}, the expectile classifier assigns $\widehat r_{\tau,d}$ to the class $k$ whose population center $r_\tau(\widetilde F_{k,d})$ is closest, using a common level $\tau$ across classes and a fixed rule for ties. Let $R_\tau(d)$ denote its misclassification probability conditional on $D=d$. We compare this classifier with the optimal classifier based on the mean aggregate, whose representation-induced Bayes risk is defined by \citep{blackwell1951comparison,reid2011information,duchi2018multiclass,sekeh2020learning}
\begin{equation}
R_{\mathrm{mean}}^*(d)
=\inf_g\Pr\{g(\overline Z_d,d)\neq Y\mid D=d\},
\nonumber
\end{equation}
where the infimum is over all measurable classifiers. Thus, $R_{\mathrm{mean}}^*(d)$ represents the minimum classification risk achievable using the mean aggregate.

\begin{theorem}[Finite neighborhood risk bound with self-inclusion]
\label{thm:shape-risk}
Under Assumption~\ref{assump:risk-fixed}, there exist explicit constants $\gamma>0$ and $B_K>0$, depending only on $\tau$, $d_0$, and the distributional parameters, such that, for every admissible finite integer $d\geq d_0$,
\begin{equation}
\frac{R_\tau(d)}{R_{\mathrm{mean}}^*(d)}
\leq
\exp\left\{-\gamma d+
\frac{\varepsilon}{\sigma_1}\sqrt d+B_K\right\}.
\label{eq:risk-fixed-dominance}
\end{equation}
In particular, $R_\tau(d)<R_{\mathrm{mean}}^*(d)$ whenever
$d>\left(\frac{\frac{\varepsilon}{\sigma_1}+\sqrt{\left(\frac{\varepsilon}{\sigma_1}\right)^2+4\gamma B_K}}{2\gamma}\right)^2$.
The explicit constants and the full proof are provided in Section~S3.1 of the Supplementary Material.
\end{theorem}

Theorem~\ref{thm:shape-risk} establishes a classification advantage of expectile aggregation. Under the stated conditions, the expectile nearest-center classifier achieves a lower risk than even the optimal classifier based solely on the mean aggregate for sufficiently large neighborhood sizes. Moreover, the theorem provides an exponentially decaying upper bound on their risk ratio as the neighborhood size increases. These results demonstrate that expectile aggregation can exploit class-dependent distributional information to achieve a provable reduction in classification error.

The underlying mechanism is straightforward. Mean aggregation primarily reflects differences in average message values, whereas expectile aggregation also responds to differences in distributional scale. When neighborhood classes are distinguished mainly by scale, the mean can discard information that the expectile retains. This result complements existing studies showing that distinguishable neighborhood distributions can support learning under heterophily \citep{ma2022homophily,wang2024understanding}. It also connects to the motivation of principal neighborhood aggregation \citep{corso2020principal}, which uses distributional statistics beyond the mean. Our result provides a classification risk justification for using an asymmetric expectile summary.

\subsubsection{Weak Neighborhood Information: A Local Efficiency Comparison}

The preceding result assumes that the neighborhood information remains fixed. We next consider a more difficult setting in which individual neighbors become progressively less informative. This regime is motivated by the low label informativeness observed in several heterophilic benchmark datasets \citep{platonov2023characterizing}. Let $a_1<\cdots<a_K$ be fixed nonzero numbers satisfying $\sum_ba_b=0$, and define
\[
H_{yb}^{(d)}=\frac1K-\frac{a_ya_b}{\sqrt d}.
\]
For each finite $d$, the classification risk is defined conditional on $D=d$, and the asymptotic analysis considers the resulting sequence of conditional neighborhood models as $d\to\infty$. For sufficiently large $d$, this defines a valid symmetric and row-stochastic mixing profile with positive entries and $H_{yy}^{(d)}<\frac{1}{K}$. As $d$ increases, the neighborhood distributions
$F_{y,d}=\sum_bH_{yb}^{(d)}G_b$ approach the same limiting mixture.

The scaling $d^{-\frac{1}{2}}$ balances the weakening class differences with the increasing number of neighbors. Consequently, the aggregate retains a nontrivial classification problem in which class separation and sampling variability are of comparable order. This construction is related to the graph separability analysis of \citet{wang2024understanding} and follows the principle of local asymptotic comparisons \citep[Chapter~14]{vandervaart1998asymptotic}. Retaining the independent self-message $Z_0\sim G_y$, we define
$\widetilde F_{y,d}=\frac{G_y+(d-1)F_{y,d}}{d}$.
Let $F_0=K^{-1}\sum_bG_b$, $m_0=\mathbb E_0Z$, and
\[
\alpha_b=\mathbb E_{G_b}|Z-m_0|,
\qquad
\beta_0=
\frac{\mathbb E_0[(Z-m_0)|Z-m_0|]}
{\operatorname{Var}_0(Z)},
\]
where $\mathbb E_0$ denotes expectation under $F_0$. These quantities characterize the amplitude information in the message distributions that can contribute to the advantage of asymmetric aggregation.

\begin{assumption}[Additional amplitude information]
\label{assump:risk-local}
In the local model above, $\sigma_b>0$ for every class, and
$\sum_{b=1}^K a_b(\alpha_b-\beta_0\mu_b)\neq0$.
\end{assumption}

Assumption~\ref{assump:risk-local} ensures that the weak neighborhood signal contains amplitude information beyond that captured by the mean, while allowing general class means and scales. When the Gaussian components share a common mean, this condition simplifies to $\sum_b a_b\sigma_b\neq0$.

For Theorem~\ref{thm:local-risk}, we compare expectile and mean aggregation using the same nearest-center classification rule, which assigns each observation to the class with the closest population center \citep{dabney2005classification,fan2008high}. Related approaches also consider asymmetric location summaries, such as quantiles, as class representatives \citep{hennig2016quantile}. With a common expectile level $\tau$ across classes, the class-$y$ population center and the corresponding classifier are defined as
\begin{equation}
c_{\tau,y}^{(d)}=r_\tau(\widetilde F_{y,d}),
\qquad
\widehat Y_{\tau,d}
=\mathop{\arg\min}_{1\leq k\leq K}
\left|\widehat r_{\tau,d}-c_{\tau,k}^{(d)}\right|,
\nonumber
\end{equation}
where ties are resolved by a fixed rule. The conditional classification risk is
\begin{equation}
R_\tau^{\mathrm{loc}}(d)
=\Pr\!\left\{\widehat Y_{\tau,d}\neq Y\mid D=d\right\}.
\label{eq:local-risk-definition}
\end{equation}
At $\tau=\frac{1}{2}$, the empirical expectile and population centers reduce to the corresponding means, yielding $R_{\mathrm{mean}}^{\mathrm{loc}}(d):=R_{\frac{1}{2}}^{\mathrm{loc}}(d)$. Here, $R_{\mathrm{mean}}^{\mathrm{loc}}(d)$ denotes the classification risk of mean aggregation under the nearest-center rule. Thus, Theorem~\ref{thm:local-risk} compares expectile and mean aggregation under an identical classification rule.

\begin{theorem}[Risk advantage under local alternatives]
\label{thm:local-risk}
Under Assumption~\ref{assump:risk-local}, there exists a fixed
$\tau\in(0,1)\setminus\{\frac{1}{2}\}$, common to all classes and independent of $d$, such that
\begin{equation}
\lim_{d\to\infty}
\left\{
R_{\mathrm{mean}}^{\mathrm{loc}}(d)
-
R_\tau^{\mathrm{loc}}(d)
\right\}>0.
\label{eq:risk-local-gap}
\end{equation}
\end{theorem}
The proof is provided in Section~S3.2 of the Supplementary Material.

Theorem~\ref{thm:local-risk} shows that expectile aggregation can retain a strict classification advantage even as the label informativeness of each individual neighbor approaches zero. Under the additional amplitude condition, a single expectile level achieves a smaller limiting classification risk than mean aggregation. This shows that expectile aggregation can retain a strict classification risk advantage even as the label informativeness of individual neighbors vanishes. The advantage arises from more efficient use of the remaining class signal. In this local regime, both the separation of aggregate class centers and their sampling variability affect classification accuracy. An appropriate expectile level improves the effective separation relative to this variability, resulting in a lower limiting risk. Thus, even when the mean retains some predictive information, expectile aggregation can extract the weak distributional signal more effectively.

Together, Theorems~\ref{thm:shape-risk} and~\ref{thm:local-risk} identify two complementary benefits of expectile aggregation. The first establishes a stronger finite neighborhood risk guarantee when class information is primarily expressed through distributional scale. The second shows that this advantage can persist when the information contributed by each neighbor progressively weakens. Both results demonstrate that the choice of aggregation statistic plays a fundamental role in converting available neighborhood information into accurate node classification.

\section{Numerical Experiments}
\label{sec:exper}
We conduct extensive experiments to evaluate the effectiveness, robustness, and practical behavior of the proposed ExpectileGCN framework. The empirical study consists of two complementary parts. First, we evaluate ExpectileGCN on a diverse collection of real-world node classification benchmarks spanning both homophilic and heterophilic graph regimes. These experiments are designed to assess whether the proposed adaptive expectile aggregation can improve predictive performance under varying neighborhood structures. Second, to further investigate robustness under unreliable graph environments, we perform controlled perturbation experiments by injecting feature noise and structural corruption into benchmark datasets. These simulations allow us to systematically examine the stability of different aggregation mechanisms when node attributes or graph topology become noisy, incomplete, or misleading. Together, these experiments aim to evaluate not only the classification accuracy of ExpectileGCN under standard settings, but also its robustness and reliability under challenging graph conditions.

\subsection{Evaluation of the Estimation Model}
To assess the effectiveness of ExpectileGCN, we compare it with a range of established graph neural network architectures. The first group consists of conventional message-passing and spectral graph neural networks, including Graph Convolutional Network (GCN, \citealp{kipf2017semi}), Graph Attention Network (GAT, \citealp{velivckovic2018graph}), Attention-based Graph Neural Network (AGNN, \citealp{thekumparampil2018attention}), Approximate Personalized Propagation of Neural Predictions (APPNP, \citealp{klicpera2019predict}), and JacobiConv \citep{wang2022JacobiConv}. The second group comprises methods employing robust aggregation, including MedianGCN \citep{chen2021median}, SoftMedoidGCN \citep{geisler2020reliable}, and GMGCN \citep{wang2023betagcn} (denoted as AGG$_p$ in the original paper). All baselines are evaluated using the same dataset splits and hyperparameter tuning protocol.
\begin{table*}[t]
	\centering
	\caption{ ~Summary statistics of benchmark datasets, including graph size, structural properties, and node feature dimensions.}
	\label{tab:dataset-stats}
	\renewcommand{\arraystretch}{0.9} 
	\setlength{\tabcolsep}{2pt}
	{\fontsize{6pt}{7pt}\selectfont
		\tabcolsep=0.12cm}
	\begin{tabular}{cccccccc}
		\Xhline{1pt}
		Dataset & Nodes & Edges & Classes & Homophily & Avg Degree & Diameter &Node Features \\
		\Xhline{0.5pt}
		pubmed         & 19,717 & 88,648  & 3  & 0.80 & 8.99  & 18   & 500  \\
		cora           & 2,708  & 13,264  & 7  & 0.81 & 9.80  & 19   & 1,433 \\
		minesweeper    & 10,000 & 39,402  & 2  & 0.68 & 7.88  & 99   & 7    \\
		questions      & 48,921 & 153,540 & 2  & 0.84 & 6.28  & 16   & 301  \\
		film           & 7,600  & 30,019  & 5  & 0.22 & 7.90  & 12   & 932  \\
		citeseer       & 2,277  & 62,742  & 6  & 0.74 & 55.11 & 28   & 3,703 \\
		texas          & 183    & 558     & 5  & 0.11 & 6.10  & 8    & 1,703 \\
		cornell        & 183    & 554     & 5  & 0.30 & 6.05  & 8    & 1,703 \\
		wisconsin      & 251    & 900     & 5  & 0.21 & 7.17  & 8    & 1,703 \\
		chameleon      & 2,277  & 62,742  & 5  & 0.23 & 55.11 & 11   & 2,325 \\
		squirrel       & 5,201  & 396,706 & 5  & 0.22 & 152.55 & 9   & 2,089 \\
		amazon-ratings & 24,492 & 93,050  & 5  & 0.38 & 7.60  & 45   & 300  \\
		roman-empire   & 22,662 & 32,927  & 18 & 0.05 & 2.91  & 6,824 & 300  \\
		\Xhline{1pt}
	\end{tabular}
\end{table*}
\subsection{Experimental Settings}
All models are trained using the Adam optimizer, with hyperparameters selected based on validation performance. For each model and dataset, we tune the hidden dimension over $\{64,128\}$, dropout rate over $\{0.3,0.5\}$, learning rate over $\{0.005,0.01\}$, and weight decay over $\{5\times10^{-4},10^{-3}\}$. Training employs early stopping, and the model with the best validation performance is used for testing. Unless otherwise specified, all models use a two-layer architecture with ReLU activation and dropout. ExpectileGCN employs a learnable expectile level in each layer, initialized at $\tau=0.5$, with smoothing parameter $s=1$ and $T=3$ fixed-point iterations. We use a $60\%/20\%/20\%$ training, validation, and test split, shared across all methods. Each experiment is repeated five times with different random seeds, and the mean and standard deviation are reported. We use cross-entropy loss for multiclass classification and binary cross-entropy loss for binary classification, reporting accuracy for multiclass datasets and ROC-AUC for minesweeper and questions. All experiments are conducted on an NVIDIA A40 GPU with 48 GB of memory.
\begin{table*}[t]
    \centering
    \renewcommand{\arraystretch}{1.1}
    \caption{Test accuracy (\%) (or ROC-AUC (\%) for minesweeper and questions) on thirteen real-world datasets. Best results are shown in bold; `OOM' denotes out-of-memory. The dashed line separates homophilic and heterophilic datasets. All-rank and Heter-rank denote average ranks over all datasets and heterophilic datasets, respectively, with lower values indicating better performance.}
    \label{tab:realworld-results}
    \resizebox{\textwidth}{!}{%
    {\fontsize{8.5pt}{9pt}\selectfont
    \tabcolsep=0.1cm
    \begin{tabular}{cccccccccc}
        \Xhline{0.5pt}
        Dataset & ExpectileGCN & GMGCN & MedianGCN & SoftMedoidGCN & GCN & GAT & APPNP & AGNN & JacobiConv \\
        \Xhline{0.5pt}
        pubmed
        & \textbf{90.12$\pm$0.21} & 89.88$\pm$0.32 & 89.34$\pm$0.12 & 88.44$\pm$0.05
        & 88.65$\pm$0.18 & 87.58$\pm$0.62 & 88.20$\pm$0.34 & 85.56$\pm$8.13 & 89.60$\pm$0.26 \\
        cora
        & 85.89$\pm$0.52 & 85.71$\pm$0.60 & 86.34$\pm$0.38 & 84.42$\pm$0.38
        & 85.75$\pm$0.19 & 85.08$\pm$0.53 & 86.85$\pm$0.19 & \textbf{87.11$\pm$0.45} & 85.75$\pm$0.47 \\
        minesweeper
        & 91.57$\pm$0.05 & 88.38$\pm$0.44 & 77.74$\pm$0.38 & 54.55$\pm$8.27
        & 66.33$\pm$8.52 & 68.41$\pm$11.59 & 58.55$\pm$1.68 & 72.96$\pm$5.22 & \textbf{93.06$\pm$1.60} \\
        questions
        & 72.78$\pm$2.13 & OOM & OOM & 35.93$\pm$0.77
        & 63.33$\pm$2.25 & 71.73$\pm$0.77 & 41.28$\pm$2.43 & 62.78$\pm$12.19 & \textbf{76.60$\pm$1.02} \\
        citeseer
        & 76.61$\pm$1.66 & 76.01$\pm$1.24 & 77.45$\pm$0.90 & \textbf{77.90$\pm$0.22}
        & 75.74$\pm$0.32 & 76.43$\pm$0.74 & 76.70$\pm$0.35 & 76.25$\pm$0.63 & 77.18$\pm$0.73 \\
        \hdashline
        film
        & \textbf{35.82$\pm$0.79} & 34.32$\pm$0.78 & 34.92$\pm$1.20 & 31.59$\pm$0.41
        & 35.03$\pm$0.77 & 33.88$\pm$0.77 & 34.63$\pm$0.65 & 34.51$\pm$0.53 & 33.39$\pm$0.15 \\
        texas
        & 82.70$\pm$7.57 & 82.70$\pm$6.53 & 81.08$\pm$4.52 & 49.73$\pm$2.16
        & 78.92$\pm$2.02 & 71.89$\pm$7.17 & 82.16$\pm$2.16 & \textbf{85.41$\pm$5.01} & 71.35$\pm$4.71 \\
        cornell
        & 71.89$\pm$4.05 & 73.51$\pm$3.15 & 74.59$\pm$2.16 & 61.08$\pm$2.16
        & 65.95$\pm$2.16 & 71.89$\pm$2.76 & 72.43$\pm$3.97 & 71.35$\pm$4.05 & 70.27$\pm$2.42 \\
        wisconsin
        & 85.88$\pm$1.47 & \textbf{87.06$\pm$3.42} & 84.71$\pm$3.80 & 57.65$\pm$2.00
        & 86.27$\pm$2.48 & 80.39$\pm$6.56 & 85.88$\pm$1.92 & 85.10$\pm$1.57 & 86.67$\pm$4.54 \\
        chameleon
        & 63.60$\pm$1.35 & 59.61$\pm$0.47 & 60.57$\pm$1.95 & 60.48$\pm$0.80
        & 63.77$\pm$1.46 & \textbf{65.66$\pm$1.08} & 54.47$\pm$0.55 & 61.89$\pm$0.87 & 57.94$\pm$1.96 \\
        squirrel
        & 46.03$\pm$1.30 & 42.31$\pm$1.12 & 44.02$\pm$0.67 & 52.55$\pm$1.14
        & 51.43$\pm$1.10 & \textbf{54.72$\pm$1.87} & 36.33$\pm$0.38 & 47.44$\pm$0.81 & 44.25$\pm$0.95 \\
        amazon-ratings
        & 53.42$\pm$0.08 & \textbf{53.72$\pm$0.53} & 52.53$\pm$0.43 & 43.58$\pm$0.21
        & 49.23$\pm$0.76 & 50.00$\pm$0.59 & 46.76$\pm$0.39 & 46.67$\pm$0.32 & 48.08$\pm$4.69 \\
        roman-empire
        & 79.41$\pm$0.22 & 78.68$\pm$0.18 & 78.56$\pm$0.43 & 41.06$\pm$0.42
        & 68.16$\pm$0.30 & 68.28$\pm$0.38 & 70.25$\pm$0.94 & 72.97$\pm$0.30 & \textbf{80.86$\pm$0.41} \\
        \Xhline{0.5pt}
        All-rank
        & \textbf{2.96} & 4.54 & 4.35 & 7.23 & 5.19 & 5.42 & 5.65 & 5.08 & 4.58 \\
        Heter-rank
        & \textbf{2.85} & 4.20 & 4.75 & 7.80 & 4.80 & 4.85 & 6.05 & 4.90 & 4.80 \\
        \Xhline{1pt}
    \end{tabular}
    }%
    }
\end{table*}
\subsection{Evaluation on Node Classification Benchmarks}
\begin{figure}[htbp]
	\centering
	\includegraphics[width=0.8\textwidth]{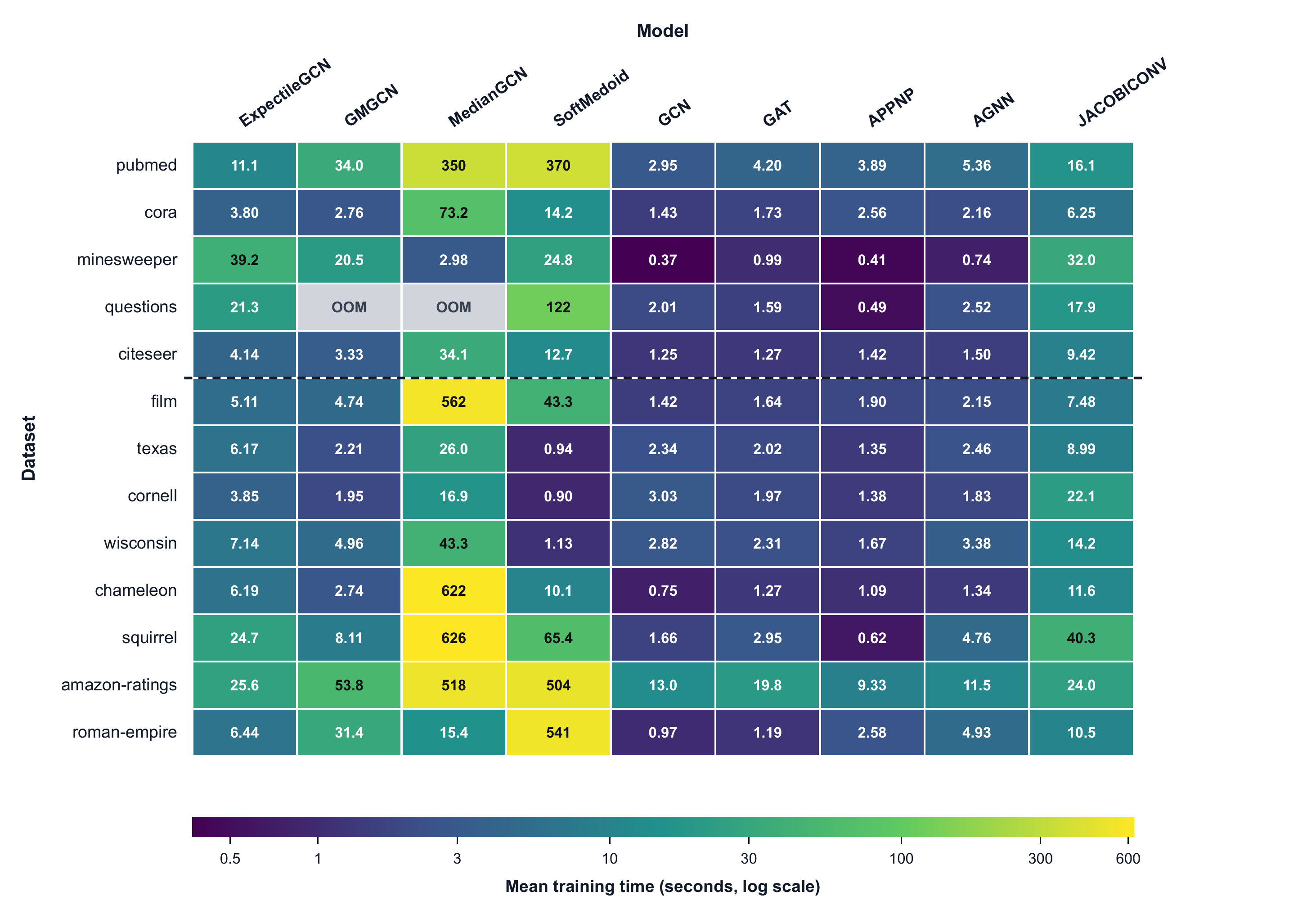}
	\caption{~Heatmap of the average training time (in seconds) for each method across different datasets. The reported time is the wall-clock training cost averaged over 5 runs.}
	\label{fig:train_time}
\end{figure}
To evaluate the predictive performance of ExpectileGCN, we conduct experiments on thirteen real-world benchmark datasets: pubmed, cora, minesweeper, questions, citeseer, film, texas, cornell, wisconsin, chameleon, squirrel, amazon-ratings, and roman-empire. These datasets cover both homophilic and heterophilic graph structures, allowing us to assess the effectiveness of expectile aggregation across different neighborhood characteristics. We follow a semi-supervised node classification setting, using the same training, validation, and test splits for all methods.
Table~\ref{tab:realworld-results} presents the classification results and average rankings across thirteen benchmark datasets. ExpectileGCN achieves the lowest average rank of 2.96 across all datasets and 2.85 across the eight heterophilic datasets, ranking first among all nine methods in both evaluations. It obtains the best results on pubmed and film and performs competitively on the remaining datasets, including minesweeper, questions, amazon-ratings, and roman-empire. Compared with other aggregation methods, ExpectileGCN achieves better average rankings than GMGCN, MedianGCN, and SoftMedoidGCN, while remaining competitive with conventional and spectral GNNs. These findings demonstrate the effectiveness of expectile aggregation across diverse graph structures, with particularly strong overall performance on heterophilic datasets.
Figure~\ref{fig:train_time} reports the average training time of each model on the benchmark datasets with early stopping. ExpectileGCN is substantially faster than MedianGCN and SoftMedoidGCN on most datasets, with particularly pronounced differences on larger graphs such as pubmed, amazon-ratings, and roman-empire. Compared with GMGCN, ExpectileGCN achieves shorter training times on these three datasets, while GMGCN is generally faster on smaller graphs. Although expectile aggregation introduces additional computational overhead relative to standard message-passing models due to its fixed-point iterations, it avoids the sorting and pairwise neighborhood operations required by MedianGCN and SoftMedoidGCN. Together with the classification results in Table~\ref{tab:realworld-results}, these findings demonstrate that ExpectileGCN offers a favorable balance between predictive performance and computational efficiency, particularly on larger graphs.
\subsection{Robustness under Feature and Structural Perturbations}
To evaluate robustness under challenging graph conditions, we conduct perturbation experiments on most of the benchmark datasets. These two datasets are excluded because the benchmark results indicate relatively low classification performance, making robustness comparisons less informative. We select six representative models covering different aggregation mechanisms: GCN, AGNN, JacobiConv, MedianGCN, GMGCN, and the proposed ExpectileGCN. These methods include conventional and spectral graph neural networks, robust median aggregation, and nonlinear adaptive aggregation, allowing us to examine how different aggregation strategies respond to perturbations. We consider both node feature corruption and graph structural perturbations. For each dataset and perturbation setting, specified by a feature noise level and a structural modification ratio, we repeat the experiments with 5 random seeds and report the mean and standard deviation of the test performance.\\
\textbf{Feature noise.} To simulate attribute corruption, we independently add Gaussian noise to each feature dimension of every node, yielding $\widetilde{\mathbf X}=\mathbf X+\boldsymbol{\epsilon}$, where $\epsilon_{ij}\sim\mathcal N(0,\sigma^2)$ and $\sigma\in\{0.5,1.0\}$. This perturbation models measurement errors and other sources of noise in node attributes.\\
\textbf{Structural perturbations.} We perturb the graph topology at levels $p\in\{0,0.2,0.4,0.6\}$ through two operations: (i) randomly removing a fraction $p$ of edges connecting nodes with the same label to reduce homophily, and (ii) adding edges between nodes with different labels, with the number of added edges equal to $p$ times the original edge count to increase heterophilic interference. These perturbations are applied separately or jointly to evaluate robustness under different structural corruption patterns.
\begin{table*}[t]
    \centering
    \caption{\footnotesize Average rank of each method across datasets.
    Methods are ranked by test performance under each of the 32
    perturbation settings, and the reported values are averaged
    across settings. Lower ranks indicate better performance.
    Best results are shown in bold. The dashed line separates
    homophilic and heterophilic datasets. All-rank and Heter-rank
    denote average ranks over all datasets and heterophilic
    datasets, respectively.}
    \label{tab:simulation-earstop-average-ranks}
    \renewcommand{\arraystretch}{1.1}
    \resizebox{\textwidth}{!}{
    {\fontsize{10pt}{10pt}\selectfont
    \begin{tabular}{ccccccc}
        \Xhline{1pt}
        Dataset & ExpectileGCN & GMGCN & JacobiConv
        & MedianGCN & GCN & AGNN \\
        \Xhline{0.5pt}
        pubmed         & \textbf{1.19} & 2.88 & 3.25 & 4.81 & 5.59 & 3.28 \\
        cora           & 1.91 & 3.03 & \textbf{1.16} & 3.91 & 5.27 & 5.73 \\
        citeseer       & 1.78 & 3.50 & \textbf{1.31} & 3.41 & 5.00 & 6.00 \\
        minesweeper    & 2.66 & 4.38 & \textbf{1.06} & 3.22 & 4.88 & 4.81 \\
        questions      & 1.84 & 5.50 & \textbf{1.72} & 5.50 & 3.44 & 3.00 \\
        \hdashline
        film           & \textbf{1.56} & 2.88 & 2.19 & 3.38 & 5.66 & 5.34 \\
        texas          & 1.95 & \textbf{1.20} & 4.38 & 4.14 & 5.34 & 3.98 \\
        cornell        & 3.66 & 4.06 & 3.02 & \textbf{2.89} & 3.92 & 3.45 \\
        wisconsin      & 1.95 & \textbf{1.17} & 4.30 & 3.22 & 4.83 & 5.53 \\
        chameleon      & \textbf{1.00} & 3.27 & 2.81 & 3.41 & 4.52 & 6.00 \\
        squirrel       & \textbf{1.00} & 3.94 & 2.75 & 2.69 & 5.03 & 5.59 \\
        amazon-ratings & \textbf{2.22} & 4.11 & 2.86 & 5.20 & 2.41 & 4.20 \\
        roman-empire   & \textbf{1.78} & 2.13 & 2.13 & 3.97 & 6.00 & 5.00 \\
        \Xhline{0.5pt}
        All-rank       & \textbf{1.88} & 3.23 & 2.53 & 3.83 & 4.76 & 4.76 \\
        Heter-rank     & \textbf{1.96} & 3.26 & 2.72 & 3.76 & 4.60 & 4.69 \\
        \Xhline{1pt}
    \end{tabular}
    }
    }
\end{table*}

Table~\ref{tab:simulation-earstop-average-ranks} summarizes the robustness rankings across different perturbation settings. ExpectileGCN achieves the lowest All-rank of $1.88$, outperforming the second-best JacobiConv ($2.53$). It also attains the best Heter-rank of $1.96$, compared with $2.72$ for JacobiConv, and ranks first on six of the thirteen datasets. These results demonstrate the strong overall robustness of ExpectileGCN across both homophilic and heterophilic graphs.

\begin{figure}[htbp]
	\centering
	\includegraphics[width=\textwidth]{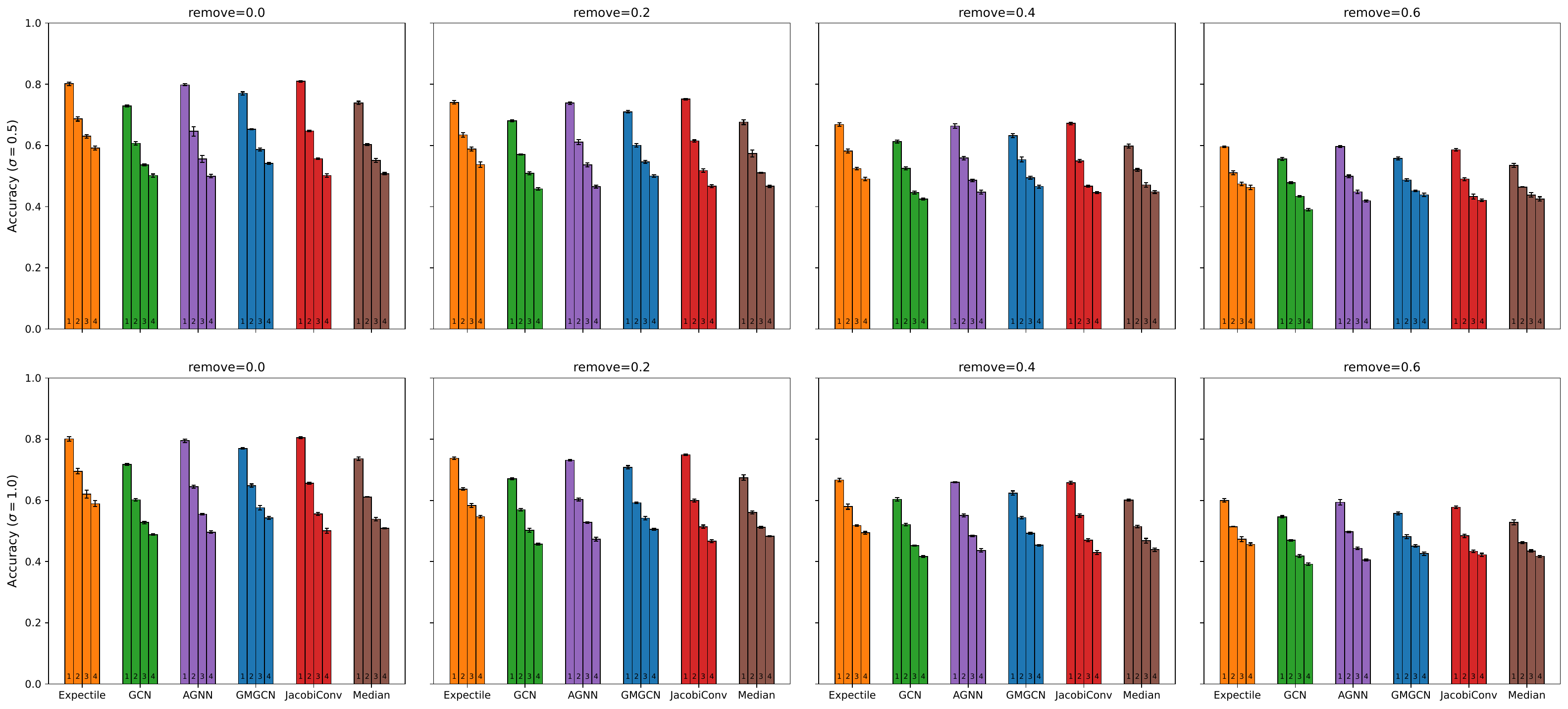}
	\caption{Robustness comparison of different GNN models on the pubmed dataset. Within each subplot, each bar shows the mean test accuracy under the corresponding added edge proportion $p=\{0,0.2,0.4,0.6\}$, marked as 1, 2, 3, and 4 below the bars. Black error bars denote the standard deviation across 5 independent runs.}
	\label{fig:pubmed-robustness}
\end{figure}

\begin{figure}[htbp]
	\centering
	\includegraphics[width=\textwidth]{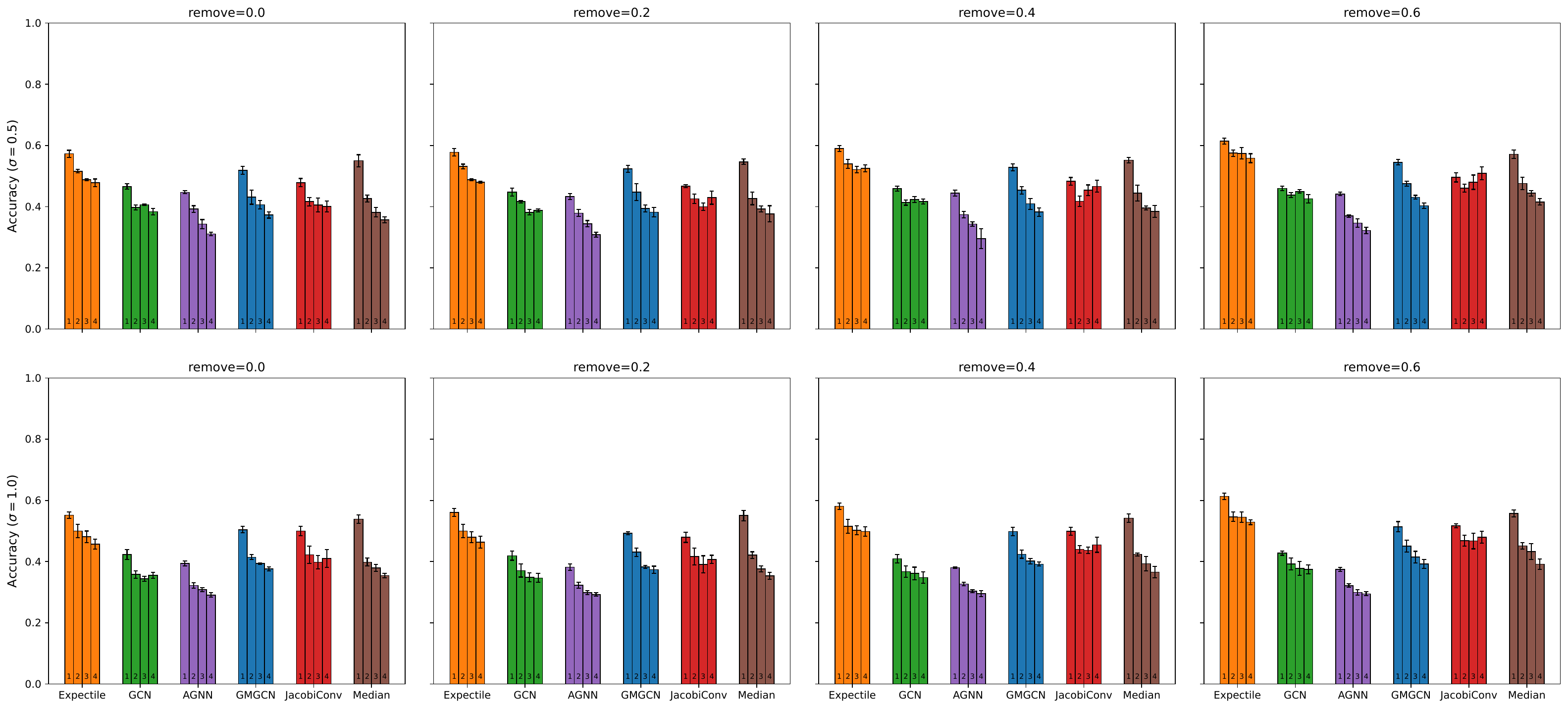}
	\caption{Robustness comparison of different GNN models on the chameleon dataset. Within each subplot, each bar shows the mean test accuracy under the corresponding added edge proportion $p=\{0,0.2,0.4,0.6\}$, marked as 1, 2, 3, and 4 below the bars. Black error bars denote the standard deviation across 5 independent runs.}
	\label{fig:chameleon-robustness}
\end{figure}

Figures~\ref{fig:pubmed-robustness} and~\ref{fig:chameleon-robustness} further illustrate robustness under feature and structural perturbations on representative homophilic and heterophilic datasets. As perturbation levels increase, ExpectileGCN generally maintains competitive performance, particularly on the heterophilic \texttt{chameleon} dataset. These findings are consistent with the theoretical insights of Theorems~\ref{thm:shape-risk} and~\ref{thm:local-risk}, which highlight how expectile aggregation can exploit class-dependent distributional information beyond the mean. Although these theorems do not directly characterize robustness under graph perturbations, they provide a complementary explanation for the potential benefits of expectile aggregation when neighborhood information is challenging to extract.

Similar robustness patterns are observed across the remaining datasets. Detailed results and visualizations for individual datasets are provided in Appendix~\ref{app:robustness}, and hyperparameter sensitivity analyses are reported in Appendix~\ref{app:sensitivity}.

\section{Conclusion}
\label{sec:con}
In this article, we propose ExpectileGCN, a graph neural network that introduces expectile aggregation to flexibly capture neighborhood information beyond the mean. By learning the expectile level, ExpectileGCN adaptively weights neighborhood messages and provides a robust alternative to conventional mean aggregation. We establish transductive generalization guarantees and stability properties for the proposed model. We further show that expectile aggregation can achieve lower classification risk than mean aggregation by exploiting informative distributional structure in neighborhoods, and that this advantage can persist when the information contributed by individual neighbors becomes weak. Extensive experiments on both homophilic and heterophilic graphs demonstrate competitive predictive performance and improved robustness under feature and structural perturbations.

We conclude with two directions for future research. First, the current model uses a common expectile level in neighborhood aggregation. Allowing the expectile level to adapt across nodes or local graph structures may provide greater flexibility for graphs with heterogeneous neighborhood distributions. Second, it would be interesting to extend expectile aggregation to more general relational settings, including dynamic graphs, hypergraphs, and other higher-order structures, and investigate the corresponding statistical and generalization properties.

\section*{Acknowledgments and Disclosure of Funding}
Wei Lan’s research was supported by the National Key R\&D Program of China (2022YFA1003702), the National Natural Science Foundation of China (No.72422020, 72333001, 12531011), the Joint Lab of Data Science and Business Intelligence at Southwestern University of Finance and Economics.

\clearpage
\appendix
\setcounter{section}{0}
\renewcommand{\thesection}{A\arabic{section}}
\numberwithin{equation}{section}
\setcounter{table}{0}
\renewcommand{\thetable}{A\arabic{table}}
\setcounter{figure}{0}
\renewcommand{\thefigure}{A\arabic{figure}}
\allowdisplaybreaks[4]
\section*{Appendix}
This appendix reports additional robustness experiments and hyperparameter sensitivity results. All proofs and auxiliary lemmas are provided in the Supplementary Material; section and lemma numbers prefixed by S refer to that document.

\section{Additional Experimental Results}
\label{app:experiments}
    \subsection{Additional Robustness Visualizations}
\label{app:robustness}
\begin{figure}[htbp]
		\centering
		\includegraphics[width=\textwidth]{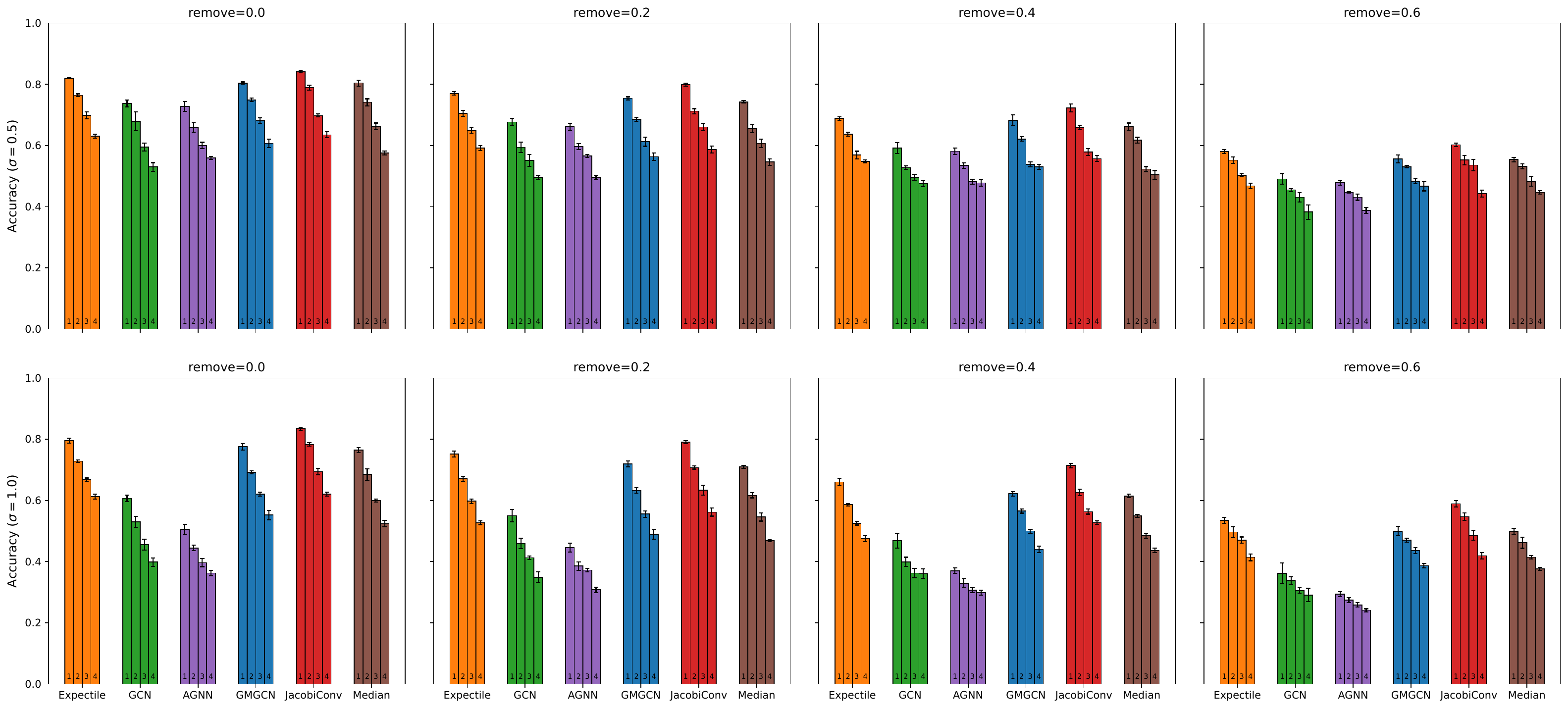}
		\caption{~Robustness comparison of different GNN models on the cora dataset. Within each subplot, each bar shows the mean test accuracy under the corresponding value of added-edge proportion
			$p=\{0, 0.2, 0.4, 0.6\}$, marked as 1, 2, 3, and 4 on the bottom of the bars.
			Black error bars denote the standard deviation across 5 independent runs.}
		\label{fig:cora-robustness}
	\end{figure}

\begin{figure}[htbp]
		\centering
		\includegraphics[width=\textwidth]{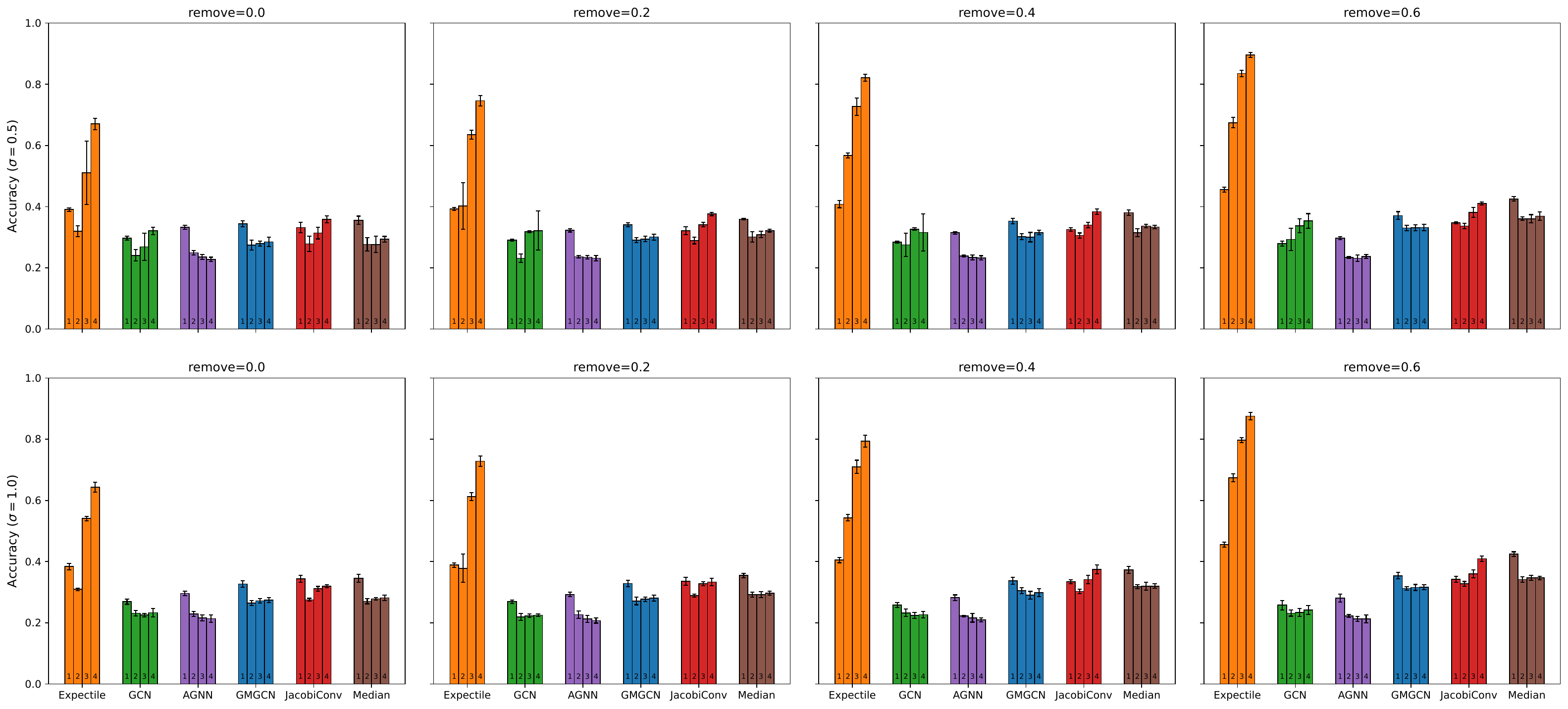}
		\caption{\footnotesize~Robustness comparison of different GNN models on the squirrel dataset. Within each subplot, each bar shows the mean test accuracy under the corresponding value of added-edge proportion
			$p=\{0, 0.2, 0.4, 0.6\}$, marked as 1, 2, 3, and 4 on the bottom of the bars.
			Black error bars denote the standard deviation across 5 independent runs.}
		\label{fig:squirrel-robustness}
	\end{figure}

\begin{figure}[htbp]
		\centering
		\includegraphics[width=\textwidth]{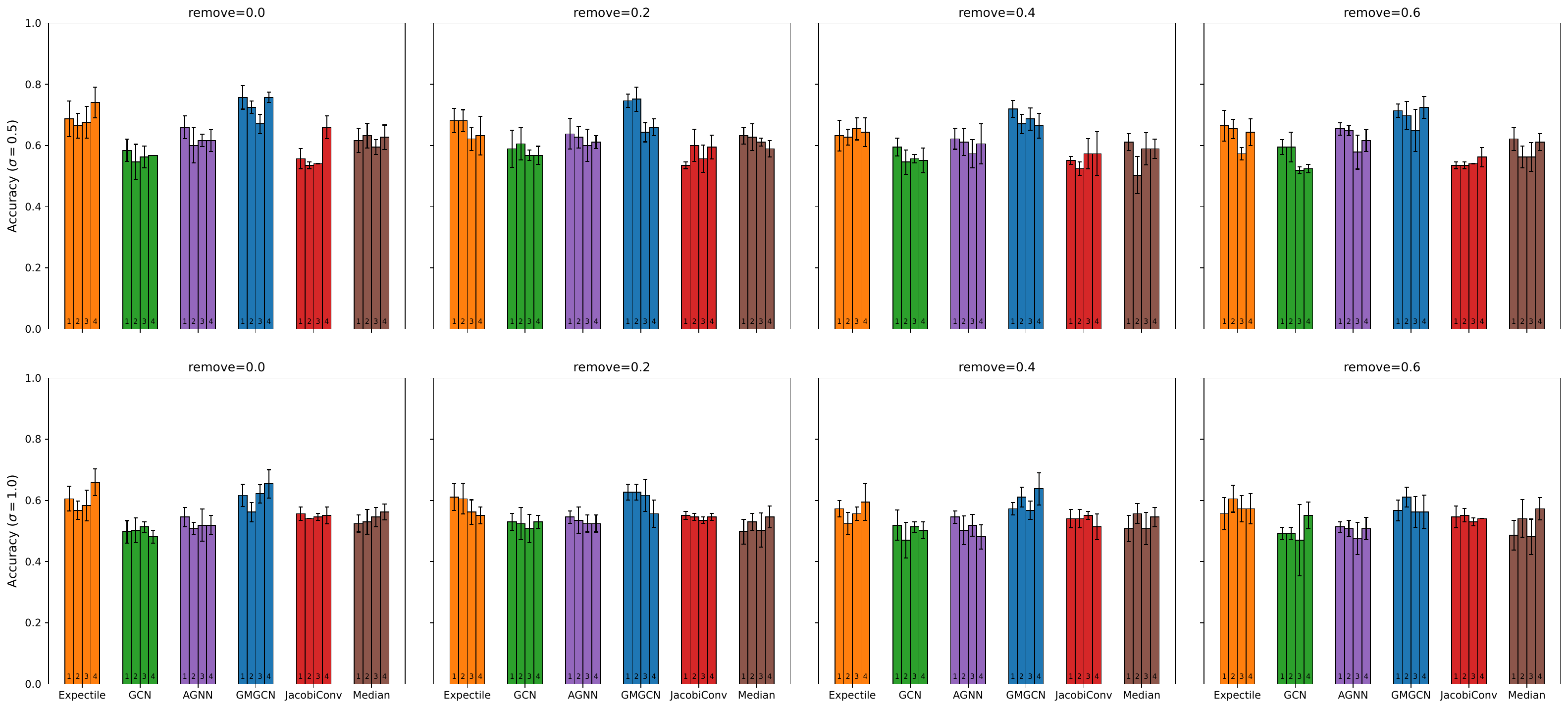}
		\caption{\footnotesize~Robustness comparison of different GNN models on the texas dataset. Within each subplot, each bar shows the mean test accuracy under the corresponding value of added-edge proportion
			$p=\{0, 0.2, 0.4, 0.6\}$, marked as 1, 2, 3, and 4 on the bottom of the bars.
			Black error bars denote the standard deviation across 5 independent runs.}
		\label{fig:teax-robustness}
	\end{figure}

\begin{figure}[htbp]
		\centering
		\includegraphics[width=\textwidth]{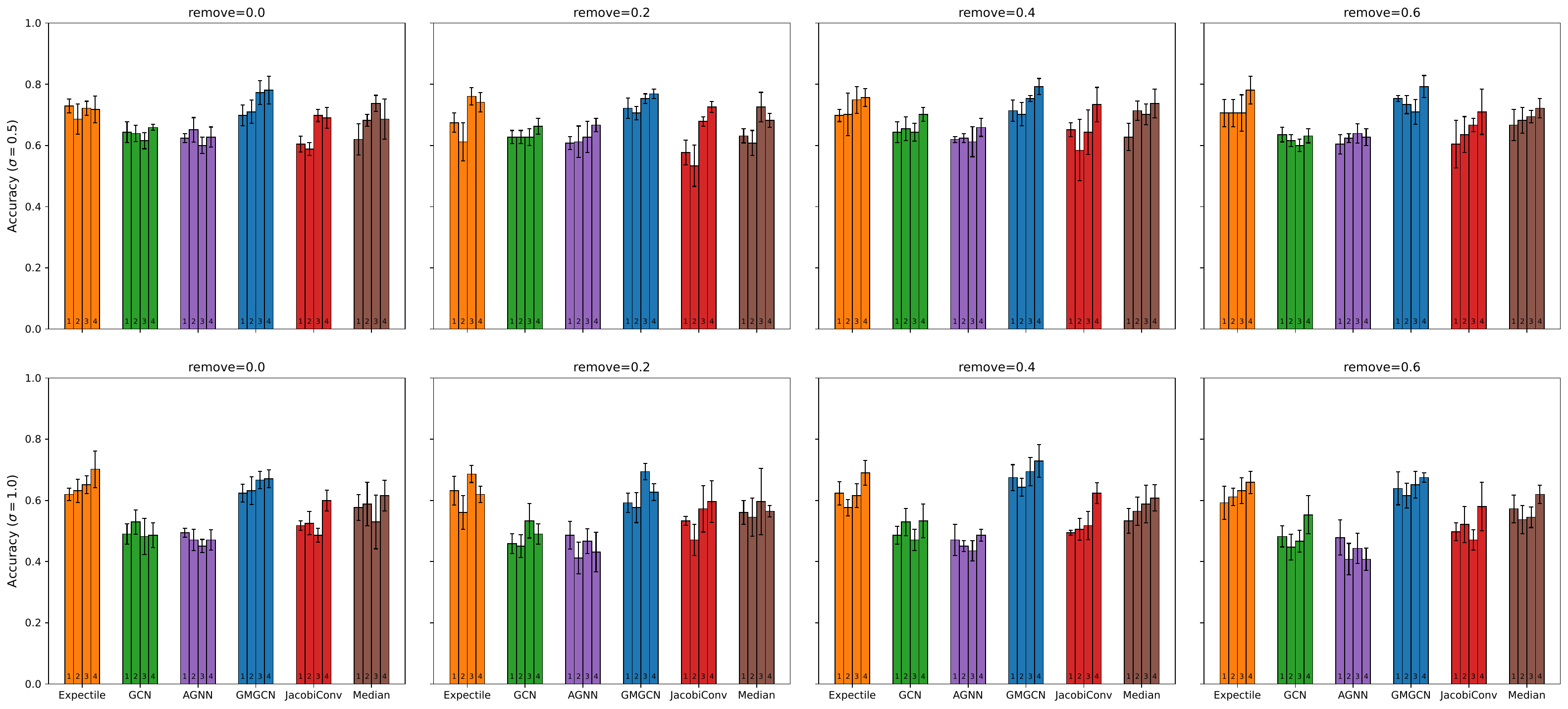}
		\caption{\footnotesize~Robustness comparison of different GNN models on the wisconsin dataset. Within each subplot, each bar shows the mean test accuracy under the corresponding value of added-edge proportion
			$p=\{0, 0.2, 0.4, 0.6\}$, marked as 1, 2, 3, and 4 on the bottom of the bars.
			Black error bars denote the standard deviation across 5 independent runs.}
		\label{fig:wisconsin-robustness}
	\end{figure}

	The robustness experiments on the other datasets follow the same perturbation protocol as described in the main text. Figures~\ref{fig:cora-robustness}--\ref{fig:wisconsin-robustness} present detailed comparisons on four datasets: cora (homophilic), and squirrel, texas, and wisconsin (heterophilic). The figures report classification performance under different levels of feature corruption and structural perturbation, allowing comparisons of the methods across both homophilic and heterophilic settings.
	
\FloatBarrier
\subsection{Hyperparameter Sensitivity Analysis}
\label{app:sensitivity}
	
We investigate the sensitivity of ExpectileGCN to two key hyperparameters: the number of fixed-point iterations $T$ and the sigmoid sharpness parameter $s$. We consider $T\in\{1,2,4,8\}$ and $s\in\{1,3,5,10\}$ across nine benchmark datasets covering both homophilic and heterophilic graphs. For each configuration, we report the mean classification accuracy and standard deviation over five independent runs.
	
\begin{table*}[htbp]
    \centering
    \renewcommand{\arraystretch}{1.2}
    \caption{\footnotesize Sensitivity analysis of ExpectileGCN with
    respect to the number of fixed-point iterations $T$ and smoothing
    sharpness $s$. Results are reported as mean accuracy (\%)
    $\pm$ standard deviation over 5 runs.}
    \label{tab:sensitivity}
    {\fontsize{8pt}{9pt}\selectfont
    \tabcolsep=0.075cm
    \begin{tabular}{c c c c c c c c c c c}
        \Xhline{1pt}
        $T$ & $s$ & pubmed & cora & film & citeseer
        & texas & cornell & wisconsin & chameleon & squirrel \\
        \Xhline{0.5pt}
        1 & 1  & 90.01$\pm$0.18 & 85.91$\pm$0.41 & 35.68$\pm$0.70 & 76.94$\pm$1.28 & 87.03$\pm$4.80 & 71.08$\pm$3.21 & 84.12$\pm$1.85 & 63.42$\pm$1.13 & 45.73$\pm$0.69 \\
          & 3  & 89.97$\pm$0.25 & 85.86$\pm$0.46 & 35.75$\pm$0.63 & 76.98$\pm$1.25 & 85.68$\pm$6.51 & 71.62$\pm$3.02 & 82.75$\pm$2.60 & 63.68$\pm$1.36 & 45.86$\pm$0.98 \\
          & 5  & 90.03$\pm$0.26 & 85.89$\pm$0.43 & 35.66$\pm$0.55 & 76.73$\pm$1.48 & 85.14$\pm$6.42 & 71.62$\pm$3.47 & 83.73$\pm$2.78 & 63.82$\pm$1.39 & 45.97$\pm$0.72 \\
          & 10 & 89.96$\pm$0.21 & 85.86$\pm$0.46 & 35.69$\pm$0.70 & 76.38$\pm$1.80 & 84.59$\pm$6.05 & 71.62$\pm$3.47 & 84.71$\pm$2.45 & 63.62$\pm$0.94 & 46.11$\pm$0.87 \\
        \Xhline{0.5pt}
        2 & 1  & 90.05$\pm$0.21 & 85.91$\pm$0.41 & 35.68$\pm$0.70 & 76.71$\pm$1.53 & 86.76$\pm$3.30 & 71.08$\pm$3.21 & 86.08$\pm$2.05 & 63.42$\pm$1.13 & 45.86$\pm$0.84 \\
          & 3  & 90.13$\pm$0.18 & 85.84$\pm$0.49 & 35.73$\pm$0.65 & 76.98$\pm$1.25 & 85.68$\pm$6.63 & 71.62$\pm$3.02 & 83.73$\pm$3.40 & 63.68$\pm$1.36 & 45.98$\pm$0.97 \\
          & 5  & 89.96$\pm$0.18 & 85.86$\pm$0.48 & 35.71$\pm$0.61 & 76.64$\pm$1.65 & 85.41$\pm$6.64 & 71.62$\pm$3.47 & 84.12$\pm$2.83 & 63.42$\pm$1.42 & 45.97$\pm$0.72 \\
          & 10 & 90.00$\pm$0.25 & 85.84$\pm$0.45 & 35.57$\pm$0.77 & 76.59$\pm$1.64 & 84.59$\pm$6.05 & 71.35$\pm$3.01 & 84.51$\pm$2.39 & 63.42$\pm$1.24 & 46.03$\pm$0.96 \\
        \Xhline{0.5pt}
        4 & 1  & 89.96$\pm$0.22 & 85.91$\pm$0.41 & 35.38$\pm$0.91 & 76.71$\pm$1.53 & 85.95$\pm$4.65 & 71.08$\pm$3.21 & 85.88$\pm$1.92 & 63.44$\pm$1.10 & 45.61$\pm$0.60 \\
          & 3  & 89.96$\pm$0.12 & 85.84$\pm$0.49 & 35.72$\pm$0.66 & 76.98$\pm$1.25 & 85.68$\pm$6.63 & 71.62$\pm$3.02 & 83.14$\pm$3.42 & 63.68$\pm$1.36 & 45.98$\pm$0.97 \\
          & 5  & 90.04$\pm$0.28 & 85.86$\pm$0.48 & 35.69$\pm$0.65 & 76.64$\pm$1.65 & 85.41$\pm$6.64 & 71.62$\pm$3.47 & 84.90$\pm$2.78 & 63.42$\pm$1.42 & 45.98$\pm$0.74 \\
          & 10 & 89.97$\pm$0.36 & 85.86$\pm$0.46 & 35.56$\pm$0.78 & 76.61$\pm$1.62 & 84.59$\pm$6.05 & 71.35$\pm$3.01 & 83.92$\pm$2.75 & 63.42$\pm$1.24 & 46.05$\pm$0.97 \\
        \Xhline{0.5pt}
        8 & 1  & 90.00$\pm$0.23 & 85.91$\pm$0.41 & 35.69$\pm$0.68 & 76.71$\pm$1.53 & 85.95$\pm$4.65 & 71.08$\pm$3.21 & 85.88$\pm$1.92 & 63.42$\pm$1.13 & 45.60$\pm$0.63 \\
          & 3  & 90.10$\pm$0.23 & 85.84$\pm$0.49 & 35.73$\pm$0.65 & 76.98$\pm$1.25 & 85.68$\pm$6.63 & 71.62$\pm$3.02 & 83.14$\pm$3.42 & 63.68$\pm$1.36 & 45.98$\pm$0.98 \\
          & 5  & 90.05$\pm$0.17 & 85.86$\pm$0.48 & 35.70$\pm$0.61 & 76.64$\pm$1.65 & 85.41$\pm$6.64 & 71.62$\pm$3.47 & 84.90$\pm$2.78 & 63.42$\pm$1.42 & 45.97$\pm$0.72 \\
          & 10 & 89.94$\pm$0.28 & 85.84$\pm$0.45 & 35.71$\pm$0.75 & 76.61$\pm$1.62 & 84.59$\pm$6.05 & 71.35$\pm$3.01 & 83.92$\pm$2.75 & 63.42$\pm$1.24 & 46.03$\pm$0.96 \\
        \Xhline{1pt}
    \end{tabular}
    }
\end{table*}

	As shown in Table~\ref{tab:sensitivity}, ExpectileGCN demonstrates stable performance across a wide range of hyperparameter choices. Variations in both mean accuracy and standard deviation remain small for different values of $T$ and $s$. In particular, small iteration numbers such as $T=1$ or $2$ already achieve performance comparable to larger iteration budgets, indicating that the implicit expectile representations require only limited iterative refinement. Similarly, moderate smoothing parameters such as $s=3$ or $5$ perform comparably to sharper approximations (e.g., $s=10$), suggesting that the sigmoid-smoothed asymmetric coefficient is not sensitive to the precise smoothness specification. Overall, these results demonstrate strong hyperparameter robustness of ExpectileGCN with respect to both iterative depth and smoothing design, making the model practically effective without extensive dataset-specific tuning.

\FloatBarrier
\bibliography{references}
\end{document}